\documentclass{article} % For LaTeX2e
\usepackage{iclr2024_conference,times}

\usepackage{amsmath,amsfonts,bm}

\def\eqref#1{equation~\ref{#1}}
\def\1{\bm{1}}

\DeclareMathAlphabet{\mathsfit}{\encodingdefault}{\sfdefault}{m}{sl}
\SetMathAlphabet{\mathsfit}{bold}{\encodingdefault}{\sfdefault}{bx}{n}

\usepackage{hyperref}
\usepackage{url}

\usepackage{hyperref}       % hyperlinks
\usepackage{url}            % simple URL typesetting
\usepackage{booktabs}       % professional-quality tables
\usepackage{amsfonts}       % blackboard math symbols
\usepackage{nicefrac}       % compact symbols for 1/2, etc.
\usepackage{microtype}      % microtypography
\usepackage{xcolor}         % colors
\usepackage{todonotes}
\usepackage{bm}
\usepackage{wrapfig}
\usepackage{physics}
\usepackage{array}
\usepackage{mathtools}
\usepackage{stmaryrd}
\usepackage{subcaption}
\usepackage{afterpage}
\usepackage{multirow}
\newcommand{\blue}[1]{#1}
\newcommand{\colorblue}{} 
\newcommand{\bluecaption}{}

\title{SineNet: Learning Temporal Dynamics in Time-Dependent Partial Differential Equations}

\newcommand{\footnotecontent}{Equal contribution. Correspondence to: Xuan Zhang \textless xuan.zhang@tamu.edu\textgreater, Jacob Helwig \\ \textless jacob.a.helwig@tamu.edu\textgreater\ and Shuiwang Ji \textless sji@tamu.edu\textgreater.}

\author{Xuan Zhang{\normalfont\textsuperscript{1}}\thanks{\footnotecontent}, 
 Jacob Helwig{\normalfont\textsuperscript{1}}\footnotemark[1], 
 Yuchao Lin{\normalfont\textsuperscript{1}}, 
 Yaochen Xie{\normalfont\textsuperscript{1}}, 
 Cong Fu{\normalfont\textsuperscript{1}}, 
 \textbf{Stephan Wojtowytsch}{\normalfont\textsuperscript{2}}\\
\textbf{\& Shuiwang Ji}\textsuperscript{1}\\
\textsuperscript{1}Department of Computer Science \& Engineering, Texas A\&M University\\
\textsuperscript{2}Department of Mathematics, University of Pittsburgh}

\iclrfinalcopy % Uncomment for camera-ready version, but NOT for submission.
\begin{document}

\maketitle

\begin{abstract}
We consider using deep neural networks to solve time-dependent partial differential equations (PDEs), where multi-scale processing is crucial for modeling complex, time-evolving dynamics. 
% to tackle the challenge of temporal shift in field elements. 
While the U-Net architecture with skip connections is commonly used by prior studies to enable multi-scale processing, our analysis shows that the need for features to evolve across layers results in temporally misaligned features in skip connections, which limits the model's performance.
To address this limitation, we propose SineNet, consisting of multiple sequentially connected U-shaped network blocks, referred to as waves. % , dubbed waves. 
In SineNet, high-resolution features are evolved progressively through multiple stages, thereby reducing the amount of misalignment within each stage.
We furthermore analyze the role of skip connections in enabling both parallel and sequential processing of multi-scale information.
Our method is rigorously tested on multiple PDE datasets, including the Navier-Stokes equations and shallow water equations, showcasing the advantages of our proposed approach over conventional U-Nets with a comparable parameter budget. We further demonstrate that increasing the number of waves in SineNet while maintaining the same number of parameters leads to a monotonically improved performance. The results highlight the effectiveness of SineNet and the potential of our approach in advancing the state-of-the-art in neural PDE solver design.
% Our code is publicly available as part of the AIRS library (\url{https://github.com/divelab/AIRS}).
Our code is available as part of AIRS (\url{https://github.com/divelab/AIRS}).
\end{abstract}

\section{Introduction}
% \todo[inline, size=\tiny]{show some data; mention energy cascade in intro; add note about padding

% Added padding. perhaps fine without energy cascade}
% \todo[inline, size=\tiny, color=pink]{Highlight two challenges in learning dynamics: \textbf{$(a)$ multi-scale information} and \textbf{$(b)$ dynamic target due to advection}.}
Partial differential equations (PDEs) describe physics from the quantum length scale (Schr\"odinger equation) to the scale of space-time (Einstein field equations) and everywhere in between. % (dynamics of physical systems using functions of derivatives. 
Formally, a PDE is an expression in which the rates of change of one or multiple quantities with respect to spatial or temporal variation are balanced.
PDEs are popular modeling tools across many scientific and engineering fields to encode force balances, including stresses that are combinations of spatial derivatives, velocity and acceleration that are time derivatives, and external forces acting on a fluid or solid. Due to their widespread use in many fields, solving PDEs using numerical methods has been studied extensively. With advances in deep learning methods, there has been a recent surge of interest in using deep learning methods for solving PDEs~\citep{raissi2019physics,kochkov2021machine,lu2021learning,stachenfeld2022learned,li2021fourier,pfaff2021learning,takamoto2022pdebench,gupta2023towards}, which is the subject of this work.

Our main focus here is on {\em fluid dynamics}, where we typically encounter two phenomena: advection and diffusion. Diffusion in isotropic media is modeled using the heat equation $(\partial_t - \Delta) \bm u = 0$, which equates the rate of change in time $\partial_t \bm u$ to the spatial Laplacian of the function $\bm u$, an expression of second-order derivatives. The fact that a first-order derivative in time is compared to a second derivative in space leads to the \textit{parabolic scaling} in which $\sqrt t$ and $x$ behave comparably. In many numerical methods such as finite element discretizations, this behavior requires discretizations for which the length-scale in time $\delta t$ is much smaller than the spatial length-scale $\delta x$: $\delta t \ll (\delta x)^2$. To avoid excessive computation time when computing in fine resolution, multi-scale models are attractive which accurately resolve both local interfaces and long-range interactions over a larger time interval per computation step. Methods which are not localized in space, \textit{e.g.}, heat kernel convolutions or spectral methods, are able to overcome the parabolic scaling.

Unsurprisingly, the situation becomes much more challenging when advection is considered. While diffusion models how heavily localized quantities spread out to larger regions over time, advection describes the transport of quantities throughout space without spreading out. In the equations of fluid dynamics, the transport term $(\bm u\cdot \nabla)\bm u$ is non-linear in $\bm u$ and famously challenging both analytically and numerically. In particular, any numerical approximation to the solution operator which propagates $\bm u$ forward in time must be non-linear to comply with the non-linearity in the equation.

For its strength in multi-scale processing, the U-Net architecture has recently become a popular choice as an inexpensive surrogate model for classical numerical methods mapping dynamics forward in time~\citep{gupta2023towards,takamoto2022pdebench,wang2020towards}. %, as a surrogate model for classical solvers.
However, because the input to the network is earlier in time than the prediction target, latent evolution occurs in the U-Net feature maps. This implies that skip connections between the down and upsampling paths of the U-Net, which are crucial for faithfully restoring high-resolution details to the output, will contain out-of-date information, and poses challenges in settings where advection plays a significant role.

In this work, we propose \textbf{SineNet}, an architecture designed to handle the challenges arising from dynamics wherein \emph{both} diffusion and advection interact.
% which tackles {\em both} the diffusion and advection problems in time-dependent PDEs. 
SineNet features a multi-stage architecture in which each stage is a U-Net block, referred to as a wave. % , dubbed as ``wave''.
By distributing the latent evolution across multiple stages, the degree of spatial misalignment between the input and target encountered by each wave is reduced, thereby alleviating the challenges associated with modeling advection.
% each wave is required to manage a smaller amount of spatial misalignment between input and target, therefore the challenge associated with advection is alleviated.
At the same time, the multi-scale processing strategy employed by each wave effectively models multi-scale phenomena arising from diffusion.
% We further analyze the role of skip connections in enabling SineNet to process information in multiple ways. 
Based on the proposed SineNet, we further analyze the role of skip connections in enabling multi-scale processing in both parallel and sequential manners.
Finally, we demonstrate the importance of selecting an appropriate padding strategy to encode boundary conditions in convolution layers, a consideration we find particularly relevant for periodic boundaries. 
% Finally, we show that SineNet can handle various boundary conditions by choosing the right padding strategy in convolution layers, which is crucial when handling periodic boundaries.
We conduct empirical evaluations of SineNet across multiple challenging PDE datasets, demonstrating consistent performance improvements over existing baseline methods. 
We perform an ablation study in which we demonstrate performance monotonically improving with number of waves for a fixed parameter budget.  % that by increasing the number of waves while maintaining the same number of parameters, the performance of SineNet monotonically improves. 
The results highlight the potential of our approach in advancing the state-of-the-art in neural PDE solver design and open 
% up 
new avenues for future research in this field.

% \section{Problem definition}
\section{Neural approximations to PDE solution operators}\label{sec:neuralSolver}
% \todo[inline, size=\tiny, color=pink]{change section title}

% PDE solution definition
Time-evolving partial differential equations describe the behavior of physical systems using partial derivatives of an unknown multivariate function 
% $\bm u(\bm{x}, t)$ of space-time where $\bm{x}\in \Omega$ and $t\in\mathbb{R}$.
of space-time $\bm u:\Omega\times \mathbb{R}\to \mathbb{R}^M$, where $\Omega$ is the spatial domain and $\mathbb{R}$ accounts for the temporal dimension.
% The output of $\bm{u}$ is a collection of physical quantities describing the state of the system, which could include scalar components such as density and pressure, or vector components such as velocity. Various such components are then stacked into an $M$-dimensional vector.
The $M$-dimensional codomain of $\bm{u}$ consists of scalar and/or vector fields such as density, pressure, or velocity describing the state of the system.
% PDE solution operator
% Given the solution  $\bm u(\cdot, t_0)$ at time $t_0$, the solution operator $\mathcal{H}: \left(\Omega \to \mathbb{R}^M\right)\times \mathbb{R}\to \left(\Omega \to \mathbb{R}^M \right)$ evolves the solution to a later time $t$ as
% \begin{equation}\label{eq:pde_operator}
%    \bm{u}(\cdot, t) = \mathcal{H}\left(\bm{u}(\cdot, t_0), t-t_0\right), \text{ for }t>t_0.
% \end{equation}
% PDE solution operator
Given the solution $\bm u(\cdot, t_0)$ at time $t_0$, our task is to learn the forward operator mapping $\bm u(\cdot, t_0)$ to the solution at a later time $t$ given by $\bm u(\cdot ,t)$. In practice, $\bm u$ is a numerical solution to the PDE, that is, discretized in space-time onto a finite grid. As opposed to neural operators~\citep{kovachki2021neural}, which aim to learn this operator independent of the resolution of the discretization, we instead focus on learning the operator for a fixed, uniform discretization. 
% % Discretization
% To solve time-dependent PDEs numerically, we discretize both spatial and temporal dimensions. Specifically, we focus on the setting where the solution is sampled at fixed time intervals and the spatial domain is sampled on a grid with a fixed resolution. 
Concretely, 
% for a PDE defined in a $d$-dimensional space, 
for a $d$-dimensional spatial domain $\Omega$,
the discretized solution at time step $t$ 
% defines a feature map 
is given by $\bm{u}_t\in \mathbb{R}^{M\times N_1\times\dots \times N_d}$ for $t = 1,\dots, T$, where $T$ is the temporal resolution and  $N_1, \dots, N_d$ are the spatial resolutions along each dimension of $\Omega$ ($d=2$ in our experiments). 
% Our task
% Our task is to learn a function to approximate the solution operator $\mathcal{H}$ by mapping the current discretized solution to the next future time step. 
% Moreover, as is typical in numerical solvers, we condition our models on multiple historical time steps. Formally, we seek to learn the mapping
% \begin{align}
%     \begin{split}
%     \mathcal{M}: \mathbb{R}^{h\times M\times N_1\times\dots \times N_d} &\to \mathbb{R}^{M \times N_1\times\dots \times N_d}\\
%      \{\bm{u}_{t-h}, \ldots, \bm{u}_{t}\} &\mapsto \bm{u}_{t+1},\ t = h,\ldots, T-1,
%      \end{split}
% \end{align}
% where $h$ is the number of historical time steps that the models condition on.
Moreover, as is typical in numerical solvers, we condition our models on $h$ historical time steps. Formally, we seek to learn the mapping
$\mathcal{M}: \mathbb{R}^{h\times M\times N_1\times\dots \times N_d} \to \mathbb{R}^{M \times N_1\times\dots \times N_d}$, with $\mathcal{M}(\bm{u}_{t-h+1}, \ldots, \bm{u}_{t}) = \bm{u}_{t+1}$, for$\ t = h,\ldots, T-1.$

\section{Learning temporal dynamics in continuum fields}

% \todo[inline, size=\tiny, color=pink]{Tentative plan: $(a)$ Multi-scale processing is important and U-Nets can be used to learn the maps. $(b)$ However there is a misalignment issue caused by latent evolution. $(c)$ We use a multi-stage design to slow down the latent evolution. $(d)$ Architecture and design choices of SineNet. $(e)$ Skip connections enable both parallel and sequential processing of multi-scale information. $(f)$ How to handle boundary conditions: zero padding for Neumann BC and circular padding for periodic BC.}

\subsection{Multi-scale learning and U-Nets}\label{sec:mslearn_unets}
% \todo[inline, size=\tiny, color=pink]{May condense a little (or integrate some sentences into intro). Add a formula for U-Nets.}

In Fourier space, low frequency modes correspond to global information, and high frequency modes to local. Convolutions can be performed in the frequency domain via pointwise multiplication of frequency modes, resulting in parallel processing of multi-scale information~\citep{gupta2023towards}. This approach is taken by Fourier Neural Operators (FNOs)~\citep{li2021fourier, tran2023factorized, li2022fourier,rahman2022u,helwig2023group}, which directly parameterize their convolution kernel in the frequency domain.

In the spatial domain, U-Nets~\citep{ronneberger2015u} are commonly used to process multi-scale information in a sequential manner.  U-Nets extract information on various scales hierarchically by performing convolutions on increasingly downsampled feature maps before invoking a symmetric upsampling path. Skip connections between corresponding downsampling and upsampling layers are established to restore high-resolution information. 
To map $\{\bm{u}_{t-h+1}, \ldots, \bm{u}_{t}\}$ to $\bm u_{t+1}$, the U-Net encodes inputs to $\bm{x}_0$ and then processes $\bm{x}_0$ to a sequence of $L$ feature maps along a downsampling path $\bm{x}_1,\ldots,\bm{x}_L$ as
\begin{equation}\label{eq:down}
    \bm{x}_0=P(\{\bm{u}_{t-h+1}, \ldots, \bm{u}_{t}\});\; \bm x_{\ell}=f_{\ell}\left(d\left(\bm x_{\ell-1}\right)\right),\ \ell=1,\ldots, L,    
\end{equation}
where $P$ is an encoder, $d$ is a downsampling function, and $f_\ell$ composes convolutions and non-linearities.
Next, the U-Net upsamples to a sequence of feature maps $\bm x_{L+1},\ldots,\bm x_{2L}$ along the upsampling path and decodes the final feature map $\bm x_{2L}$ to $\bm u_{t+1}$ as 
\begin{equation}\label{eq:up}
    \bm x_{\ell}= g_{\ell}\left(v\left(\bm x_{\ell-1}\right), \bm x_{2L-\ell}\right),\ \ell=L+1,\ldots, 2L;\; \bm u_{t+1}=Q\left(\bm x_{2L}\right),    
\end{equation}
where $Q$ is the decoder, $v$ is an upsampling function, and $g_\ell$ concatenates the upsampled feature map with the skip connected downsampled feature map before applying convolutions and non-linearities. 

\begin{figure}[t]
  \centering
  % \todo[inline,size=\tiny]{Add corrected feature maps to fig 1}{\includegraphics[width=0.92\textwidth]{figure/misalignment_outl_new.pdf}}
  {\includegraphics[width=\textwidth]{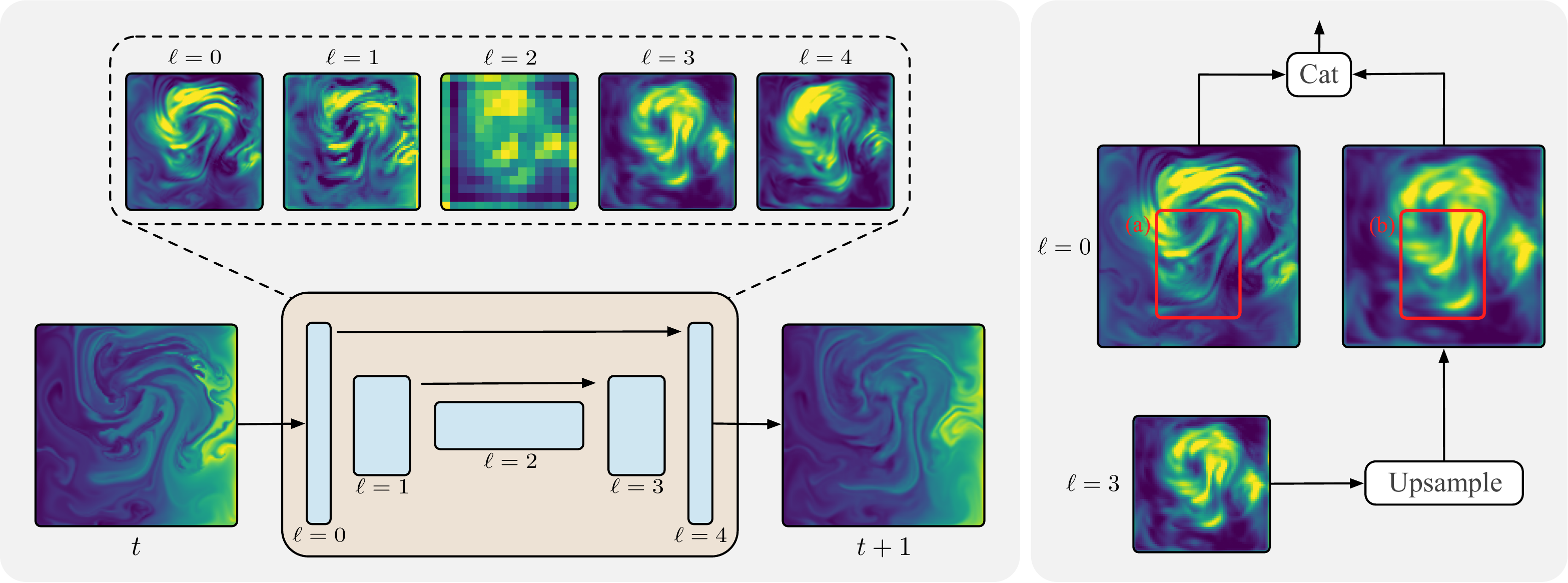}}
  \caption{Illustration of the misalignment issue in U-Nets.
  The left side show% case
s the one-step prediction of a trained U-Net, with each intermediate layer's feature maps averaged over the channel dimension displayed in the top row. 
  On the right side, the misalignment between $\ell=0$ and $\ell=3$ is demonstrated. Specifically, the feature map of $\ell=3$ is upsampled and then concatenated with $\ell=0$. The time-evolving effect of the preceding U-Net layers results in a misalignment of the corresponding physical features, as indicated by the bounding boxes (a) and (b). This misalignment is particularly problematic for convolutions. Since the kernel is localized, information from misaligned high frequency features, such as those visualized in (a) and (b), cannot be optimally integrated in updating the feature map. Mitigating this misalignment is key for improving the performance of U-Nets.}
    \vspace{-5pt} % 5
  \label{fig: misalignment}
\end{figure}

\subsection{Misalignments in learning temporal dynamics}\label{sec:misalign}
% \todo[inline, size=\tiny, color=pink]{May reduce length.}

U-Nets capture global context by aggregating localized features with downsampling layers. Using this global context, the output is constructed with upsampling layers, wherein skip connections to the downsampled feature maps play a key role in restoring high-resolution details. While we consider their application as neural solvers, U-Nets were originally developed for image segmentation~\citep{ronneberger2015u}. In segmentation tasks, features in the input image and target segmentation mask are spatially aligned. 
By contrast, in the context of learning temporal dynamics, the spatial localization property no longer holds, as temporal evolution results in \textit{misalignment} between the input and  prediction target. Here, misalignment refers to the displacement of local patterns between two feature maps, which occurs between the PDE solutions at two consecutive time steps due to advection. %}
% When mapping from the solution at one time step to the next time step with a multi-layer neural network, due to }
To resolve this misalignment, feature maps undergo latent evolution. However, due to the local processing of convolution layers, feature maps closer to the input (or output) will be more aligned with the input (or output), respectively. This leads to misalignment between the feature maps propagated via skip connections and those obtained from the upsampling path. 

To empirically demonstrate this inconsistency, we visualize the feature maps averaged over the channel dimension at each block within a well-trained U-Net in Figure~\ref{fig: misalignment}. Specifically, we examine the feature map from the skip connection at the highest resolution ($\ell=0$) and compare it with the corresponding one from the upsampling path ($\ell=3$). We observe misaligned features between $\ell=0$ and the upsampled $\ell=3$, as highlighted by the features contained in the bounding boxes (a) and (b). As kernels are most often local, \textit{e.g.}, $3\times3$, following the concatenation, convolution operations will be either partially or fully unable to incorporate the evolved features with the corresponding features in skip connections as a result of the misalignment.

\begin{figure}[t]
  \centering
  {\includegraphics[width=\textwidth]{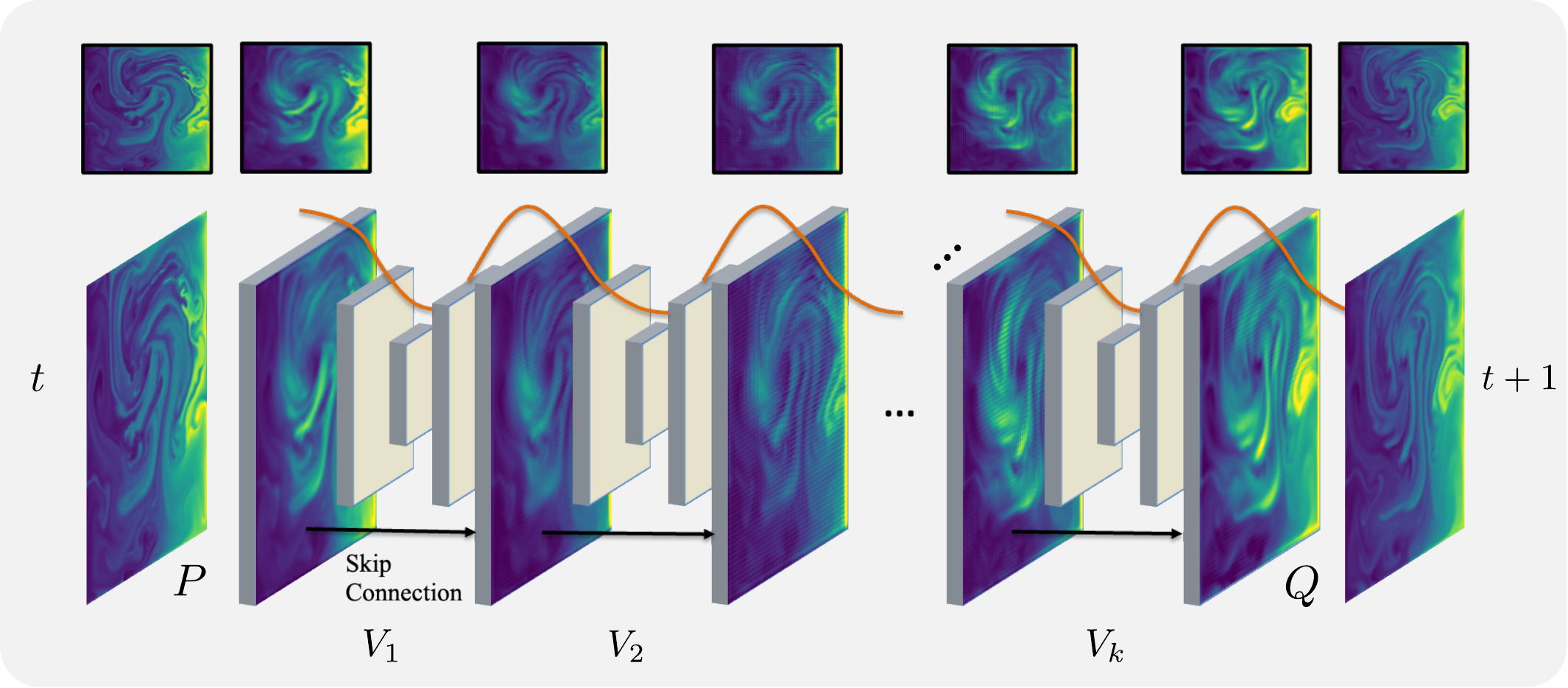}}
      \vspace{-15pt} % -7
  \caption{Illustration of the proposed SineNet for learning temporal dynamics in PDEs. Multiple U-Net waves are composed to perform one-step prediction, with the output of each wave averaged over the channel dimension displayed in the top row, demonstrating the time-evolving process from $t$ to $t+1$. The orange sinusoidal line illustrates propagation between resolutions and is not part of the model architecture. Feature maps in this figure are from \textsc{SineNet} without wave residuals for clarity and with transposed convolutions for upsampling. In Appendix~\ref{app:fmvis}, we visualize the feature maps from the \textsc{SineNet-8} presented in Section~\ref{sec:results}}
    \vspace{-10pt} % -7
  \label{fig: sinenet}
\end{figure}
% \todo[inline,size=\tiny]{Add app ref to fig 2}
\subsection{Multi-stage modeling of temporal dynamics with SineNet}\label{sec:arch}

Despite the inconsistency, skip connections in U-Nets remain crucial to restoring high-frequency details and thereby enable multi-scale computation via downsampling and upsampling. To mitigate the misalignment issue while maintaining the benefits of the U-Net architecture, we construct the \textsc{SineNet-$K$} architecture by partitioning the latent evolution across $K$ lightweight U-Nets $V_k$, referred to as waves. This approach reduces the latent evolution between consecutive downsampling and upsampling paths, which in turn reduces the degree of misalignment in skip connections. Each wave $V_k$ is responsible for advancing the system's latent state by an interval of $\delta_k\in [0,1]$, with $\sum_{1\le k\le K}\delta_k=1$. Specifically, $V_k$'s objective is to evolve the latent solution from time $t+\Delta_{k-1}$, achieved through the temporal evolution by the preceding $k-1$ waves, to time $t+\Delta_k$, where $\Delta_k$ represents the cumulative sum of all $\delta_j$ up to the $k$-th interval. The overall mapping is then given by
% Crucially, this novel configuration is designed not to require additional parameters, maintaining the efficiency of the model. 
% Building on this foundation, we introduce a simple yet effective neural architecture dubbed the SineNet-$K$ operator as shown in Figure~\ref{fig: sinenet}. This operator comprises a sequence of $K$ distinct sub-operators. Each wave $V_k$, indexed by $k=1,\cdots, K$, is responsible for advancing the system's state by an interval of $\delta_k\in [0,1]$, with $\sum_{1\le k\le K}\delta_k=1$. In this architecture, the $k$-th wave's objective is to evolve the solution from time $t+\Delta_{k-1}$, achieved through the temporal evolution by the preceding $k-1$ waves, to time $t+\Delta_k$, where $\Delta_k$ represents the cumulative sum of all $\delta_j$ up to the $k$-th interval with $\Delta_K=1$. Practically, the computation of waves is performed on the latent of the system's states. Concretely, we have 
\begin{equation}
    % \bm{u}_{t+\Delta_k} = V_k(\bm{u}_{t+\Delta_{k-1}}), \;\forall 2\le k\le K;\quad \bm{u}_{t+\Delta_1} = V_1(\{\bm{u}_{t-h+1},\cdots,\bm{u}_t\}).
    \bm{x}_{t} = P\left(\{\bm{u}_{t-h+1},\ldots,\bm{u}_t\}\right);\quad \bm{u}_{t+1} = Q\left(\bm{x}_{t+1}\right);\\
\end{equation}
\begin{equation}\label{eq:wave}
    \bm{x}_{t+\Delta_k} = V_k(\bm{x}_{t+\Delta_{k-1}}), \;k=1,\ldots,K,
\end{equation}
where $\bm{x}_{t+\Delta_k}$ denotes the latent map of $\bm{u}_{t+\Delta_k}$ at time $t+\Delta_k$, and $P$ and $Q$ are 3$\times$3 convolution layers to linearly encode and decode the solutions into or from latent maps, respectively. During training, the intermediate sub-steps as determined by $\delta_k$ are implicit and optimized end-to-end with the $K$ waves using pairs of data $\{(\{\bm{u}^j_{t-h+1},\ldots,\bm{u}^{j}_t\}, \bm u^j_{t+1})\}_{j,t}$ to minimize the objective
\begin{equation}\label{eq:loss}
    \mathbb{E}_{j,t}\left[\mathcal L\left( W_K(\{\bm{u}^j_{t-h+1},\ldots,\bm{u}^j_t\}), \bm{u}^j_{t+1}\right)\right]; \; W_K:=Q\circ V_K\circ\cdots\circ V_1\circ P,
\end{equation}
where $\mathcal L$ is a suitably chosen loss and $j$ indexes the solutions in the training set.
% analysis
% \textbf{Adaptable temporal resolution.} 
It is worth noting that the intervals $\delta_k$ are not necessarily evenly divided. They can be input-dependent and optimized for superior performance during training to enable \textbf{adaptable temporal resolution} in the latent evolution. For example, when the temporal evolution involves acceleration, it is generally more effective to use smaller time steps for temporal windows with larger velocities. We demonstrate a connection between this formulation and Neural ODE~\citep{chen2018neural} in Appendix~\ref{sec:node}.%} 

The proposed approach effectively reduces the time interval each wave is tasked with managing, thereby simplifying the learning task and mitigating the extent of misalignment in skip connections. 
%\blue{
We empirically demonstrate such effect in Appendix~\ref{sec:slow}. This strategy not only fosters a more efficient learning process but also significantly enhances the model's ability to handle complex, time-evolving PDEs. Through this approach, we present a robust way to reconcile the essential role of skip connections with the challenges they pose in the context of PDEs.

\subsection{Wave architecture}
% \todo[inline, size=\tiny, color=pink]{Give a formula for the computations. Could include a (wrap) figure for down sampling and up sampling.}

Here we discuss the construction of each of the $K$ waves $V_k$ in SineNet mapping the latent solution $\bm{x}_{t+\Delta_{k-1}}$ to $\bm{x}_{t+\Delta_k}$ in Equation~\ref{eq:wave}. $V_k$ is implemented as a conventional U-Net $U_k$, as described in Equations~\ref{eq:down} and~\ref{eq:up}, with a wave residual as
\begin{equation}\label{eq:wave_res}
V_k\left(\bm{x}_{t+\Delta_{k-1}}\right)=\bm{x}_{t+\Delta_{k-1}} + U_k(\bm{x}_{t+\Delta_{k-1}}).
\end{equation}
Each wave is constructed with a downsampling and upsampling path, both of length $L=4$,
% \emph{e.g.}
 \emph{i.e.}, $\bm x_{\ell}=f_{\ell}\left(d\left(\bm x_{\ell-1}\right)\right),\ \ell\in\{1, 2, 3, 4\}$ for downsampling and 
$\bm x_{\ell}= g_{\ell}\left(v\left(\bm x_{\ell-1}\right), \bm x_{8-\ell}\right),\ \ell\in\{5,6,7,8\}$ for upsampling.
To maintain a light-weight architecture, the downsampling and upsampling functions $d$ and $v$ are chosen as average pooling and bicubic interpolation. $f_\ell^k$ is constructed with a convolution block $c_\ell^k$, that is, two $3\times 3$ convolutions with layer norm~\citep{ba2016layer} and GeLU activation~\citep{hendrycks2016gaussian}, with the first convolution increasing the number of channels in the feature map by a multiplier $m_K$, which is chosen dependent on the number of waves $K$ in the SineNet such that the number of parameters is roughly constant in $K$. $g_\ell^k$ is constructed similarly, except the first convolution \textit{decreases} the number of channels by a factor of $m_K$. Both $f_\ell^k$ and $g_\ell^k$ additionally include a \textit{block residual}, whose role we discuss further in Section~\ref{sec:dual}, as
\begin{align}
    \label{eq:bres_down}
    f_{\ell}^k\left(d\left(\bm x\right)\right) & = w^k_\ell\left(d\left(\bm x\right)\right) + c_\ell^k\left(d\left(\bm x\right)\right)
    \\
    \label{eq:bres_up}
    g^k_{\ell}\left(v\left(\bm x\right), \bm y\right) & = w_\ell^k\left(v\left(\bm x\right)\right) + c_\ell^k\left(\mbox{cat}\left(v\left(\bm x\right), \bm y\right)\right),    
\end{align}
where $w_\ell^k$ is a point-wise linear transformation acting to increase or decrease the number of channels in the residual connection.

\subsection{Dual multi-scale processing mechanisms}\label{sec:dual}
% \todo[inline,size=\tiny]{Add disentanglement}
    While in Equation~\ref{eq:wave_res}, wave residuals serve to improve the optimization of SineNet~\citep{he2016deep}, the block residuals utilized in Equation~\ref{eq:bres_down} additionally allow each wave $V_k$ to process multi-scale information following both the sequential and parallel paradigms analyzed by~\citet{gupta2023towards}. Parallel processing mechanisms, of which the Fourier convolutions utilized by FNOs~\citep{li2021fourier} are an example, process features \textit{directly from the input} at various spatial scales independently of one another. In contrast, under the sequential paradigm inherent to U-Nets, features at each scale are extracted from the features extracted at the previous scale. Because this approach does not include a direct path to the input features, processing capacity at a given scale may be partially devoted toward maintaining information from the input for processing at later scales rather than extracting or evolving features. 
    
    To improve the efficiency of the sequential paradigm, SineNet utilizes block residuals as described in Equations~\ref{eq:bres_down} and~\ref{eq:bres_up}, which we now show allows both parallel and sequential branches of processing. Consider the input $\bm x_0$, a linear downsampling function $d$ such as average pooling, and convolution blocks $c_1, c_2, c_3$, all of which are composed as the downsampling path of a U-Net, that is, downsampling followed by convolution blocks. Assume for simplicity that the input and output channels for each of the convolution blocks are equal and $w_\ell^k$ is omitted. Then, by including block residuals, the feature map following the first downsampling and convolution block is given by
    \begin{equation}
        \bm x_1 = d\left(\bm x_0\right) + c_1\left( d\left(\bm x_0\right)\right),
    \end{equation}
    and thus, the input to $c_2$ is given by
    \begin{equation}
        d\left(\bm x_1\right) = d^2\left(\bm x_0\right) + d\left(c_1\left( d\left(\bm x_0\right)\right)\right),
    \end{equation}
    where $d^2\left(\bm x_0\right)\coloneqq d\left(d\left(\bm x_0\right)\right)$ represents the direct flow of information from the input in the parallel branch, enabling $c_2$ to process information following both the parallel and sequential paradigms. From the input to $c_3$, given by
    \begin{equation}\label{eq:disent}
        d\left(\bm x_2\right) = d^3\left(\bm x_0\right)+d\left(d\left(c_1\left( d\left(\bm x_0\right)\right)\right) + c_2\left(d\left(\bm x_1\right)\right)\right),
    \end{equation}
    it can be seen that this dual-branch framework is maintained in later layers. Specifically, the input to the $k$-th layer will be comprised of $\bm x_0$ downsampled $k$ times summed with a feature map processed sequentially. However, since addition will entangle the parallel and sequential branch, we propose to disentangle the parallel branch by concatenating it to the input of each convolution block in the downsampling path. Therefore, the downsampling path in SineNet replaces Equation~\ref{eq:bres_down} with
    \begin{equation}                    
        f_{\ell}^k\left(d\left(\bm x\right)\right) = w^k_\ell\left(d\left(\bm x\right)\right) + c_\ell^k\left(\mbox{cat}\left(d\left(\bm x\right),d^{k}\left(\bm x_0\right)\right)\right),
    \end{equation}
    where $d^{k}\left(\bm x_0\right)$ is the result of the projection layer $P$ downsampled $k$ times. We note that although latent evolution in the sequential branch will result in misalignment with the parallel branch, we empirically observe a performance gain by including the parallel branch, particularly once it has been disentangled. Additionally, this inconsistency between branches is mitigated by the multi-stage processing strategy adopted by SineNet. 

    % In general, the input to the $j$-th layer will include $\bm x$ downsampled $j-1$ times representing the parallel branch prior to processing and summed with a feature map processed sequentially. The parallel processing branch serves to maintain global features of the input relevant to subsequent layers that convolution blocks may expend unnecessary capacity preserving. Block residuals instead permit all operations to be devoted toward evolving the latent field. Although the evolution in these convolutions will result in a spatial misalignment of the maps in the parallel branch and the sequential branch, this inconsistency is again mitigated by the multi-stage processing strategy adopted by SineNet. 

\subsection{Encoding boundary conditions}\label{sec:bound}
% \todo[inline, size=\tiny, color=pink]{Briefly mention why different boundary conditions are used. We use zero padding for zero (Dirichlet for velocity and Neumann for scalar) BC and circular padding for periodic BC.}

    The boundary conditions of a PDE determine the behavior of the field along the boundary. Our experiments with the incompressible Navier-Stokes equations use a Dirichlet boundary condition on the velocity field such that the velocity on the boundary is zero, and a Neumann boundary condition on the scalar particle concentration such that the spatial derivative along the boundary is zero. These conditions are encoded in feature maps using standard zero padding. 

    The remaining PDEs we consider have periodic boundary conditions, wherein points on opposite boundaries are identified with one another such that the field ``wraps around". For example, for a function $\bm u$ on the unit torus, that is, a periodic function with domain $[0,1]^2$, we have that for $x\in[0,1]$, $\bm u(x,0)=\bm u(x,1)$ and $\bm u(0,x)=\bm u(1,x)$. For such a boundary condition, we found circular padding to be a simple yet crucial component for achieving optimal performance~\citep{dresdner2023learning}, as otherwise, a great deal of model capacity will be spent towards sharing information between two boundary points that appear spatially distant, but are actually immediately adjacent due to periodicity. As Fourier convolutions implicitly assume periodicity, FNO-type architectures are are ideally suited for such PDEs.

\section{Related work}
% \todo[inline, size=\tiny, color=pink]{Discuss FNO here. May reuse some text from the previous 3.4 (in appendix now). Could mention diffusion very briefly (in appendix now)}

\textbf{Neural PDE solvers.} Many recent studies explore solving PDEs using neural networks and are often applied to solve time-dependent PDEs~\citep{poli2022transform, lienen2022learning}. Physics-informed neural networks (PINNs)~\citep{raissi2019physics, wang2021learning, li2021physics} and hybrid solvers~\citep{um2020solver, kochkov2021machine, Holl2020Learning} share similar philosophies with classical solvers, where the former directly minimizes PDE objectives and the latter learns to improve the accuracy of classical solvers.
On the other hand, many works focus on purely data-driven learning of mappings between PDE solutions, which is the approach we take here. 
\citet{wang2020towards} and \citet{stachenfeld2022learned} apply their convolutional architectures to modeling turbulent flows in $2$ spatial dimensions, while \citet{lienen2023generative} use diffusion models to simulate $3$-dimensional turbulence. \citet{li2021fourier} developed the FNO architecture which has been the subject of several follow-up works~\citep{poli2022transform,tran2023factorized,helwig2023group}. 
Similar to SineNet, PDE-Refiner~\citep{lippe2023pde} solves PDEs via iterative application of a U-Net, however, instead of advancing time with each forward pass, U-Net applications after the first serve to refine the prediction.
% For modeling dynamics on irregular discretizations, graph neural networks have been used ~\citep{brandstetter2022message,pfaff2021learning,wu2022learning}.
% Please refer to section 9 of~\citet{zhang2023artificial} for a more detailed review of neural PDE solvers.

\textbf{Stacked U-Nets.} There are other studies exploring stacked U-Nets, however, most focus on computer vision tasks such as image segmentation~\citep{xia2017w, shah2018sunets, zhuang2018laddernet, fu2019stacked} and human pose estimation~\citep{newell2016stacked}. There are also researchers exploring U-Nets and variants therein to model PDE solutions. \citet{chen2019u} apply variations of U-Nets, including stacked U-Nets, to predict the steady state of a fluid flow, although they do not study temporal evolution and furthermore do not consider a stack greater than 2. \citet{raonic2023convolutional} propose to use CNNs to solve PDEs, but they do not have the repeated downsampling and upsampling wave structure. We differ from these works by making the key observation of feature evolution and feature alignment.

\section{Experiments}
    % \todo[inline,size=\tiny]{References to appendix; add code link!}
        As fluid dynamics are described by time-evolving PDEs where diffusion and advection play major roles, we perform experiments on multiple fluid dynamics datasets derived from various forms of the Navier-Stokes equation. We begin by describing our experimental setup, datasets and models in Sections~{\ref{sec:setup}-\ref{sec:models}} before presenting our primary results in Section~\ref{sec:results}. We close with an abalation study in Section~\ref{sec:ablation}. Furthermore, in Appendix~\ref{sec:slow}, we conduct an experiment to validate our claim of reduced latent evolution per wave, and demonstrate that this results in reduced misalignment in skip connections in Appendix~\ref{app:fmvis}. 
        % We release our code in an anonymous GitHub which will be made public following the review period (\href{https://anonymous.4open.science/r/SineNet-03B2/}{https://anonymous.4open.science/r/SineNet-03B2/}).

    \subsection{Setup and datasets}\label{sec:setup}

        In all experiments, models are trained with data pairs $\left(\{\bm{u}_{t-h+1}, \dots, \bm{u}_{t}\}, \bm{u}_{t+1}\right)$ where the inputs are the fields at current and historical steps and the target is the field at the next time step. During validation and testing, models perform an autoregressive rollout for several time steps into the future. 
        % \todo[inline,size=\tiny]{add details of scaled L2 to app}
        Scaled $L_2$ loss~\citep{gupta2023towards, tran2023factorized, li2021fourier} is used as training loss and evaluation metric, which we describe in Appendix~\ref{app:loss}. We report both the one-step and rollout test errors.
        
    % \subsection{Datasets}\label{sec:data}
        % \blue{
        We consider three datasets in our experiments. Each dataset consists of numerical solutions for a given time-evolving PDE in two spatial dimensions with randomly sampled initial conditions. Detailed dataset descriptions can be found in Appendix~\ref{sc:dataset_details}.%}
        % Vreugdenhil: This does not automatically mean that the fluid density is constant, but rather that it is independent of pressure p. The density may still vary due to other reasons, such as variations of temperature or salinity.
        
        \textbf{Incompressible Navier-Stokes (INS)}. The incompressible Navier-Stokes equations model the flow of a fluid wherein the density is assumed to be independent of the pressure, but may not be constant due to properties of the fluid such as salinity or temperature~\citep{vreugdenhil1994numerical}. The equations are given by 
        \begin{equation}\label{eq:ns}
            \frac{\partial \bm{v}}{\partial t} = -\bm{v}\cdot \nabla \bm{v} + \mu \nabla^2 \bm{v} - \nabla p + \bm{f},\ \nabla \cdot \bm{v} = 0,
        \end{equation}
        where $\bm{v}$ is velocity, $p$ is internal pressure, and $\bm{f}$ is an external force. We use the dataset from~\citet{gupta2023towards}, simulated with a numerical solver from the $\Phi_{\mbox{\tiny{Flow}}}$ package~\citep{holl2020phiflow}. In Appendix~\ref{app:conditional}, we present results on the conditional version of this dataset considered by \cite{gupta2023towards}, where the task is to generalize over different time step sizes and forcing terms.%}
         % Each trajectory is 14 time steps spaced 1.5 seconds apart and models the evolution of a vector velocity field with a Dirichlet boundary condition and a scalar field with a Neumann boundary conditon representing the concentration of particles advected by the velocity field. Trajectories are spatially discretized onto a $128\times 128$ grid. The data follow a train/valid/test split of 5,200/1,300/1,300, and the length of the time history $h$ is 4 steps such that trajectories are unrolled for 10 steps during evaluation. The viscosity $\mu$ controlling the level of turbulence in the flow is $0.01$.

        % Anderson: The Mach number is the ratio of the flow velocity to the speed of sound
        % Anderson: A flow in which the density ρ is constant is called incompressible. In contrast, a flow where the density is variable is called compressible.
        \textbf{Compressible Navier-Stokes (CNS)}. Compressibility is generally a consideration most relevant for fast-moving fluids~\citep{anderson2011ebook}. We generate our CNS dataset using the numerical solver from~\citet{takamoto2022pdebench}. 
        % Each trajectory has 21 time steps and consists of a scalar pressure field, a scalar density field, and a vector velocity field, each with periodic boundary conditions. Trajectories are generated on a $512\times 512$ spatial grid and downsampled to $128\times 128$. The dataset is split as 5,400/1,300/1,300 and, following~\citet{takamoto2022pdebench}, models use time history $h=10$ such that trajectories are unrolled for $11$ steps during evaluation. 
        The dynamics in this data are more turbulent than those in the INS data, with the viscosity as $1\times10^{-8}$ and the initial Mach number, which quantifies the ratio of the flow velocity to the speed of sound in the fluid~\citep{anderson2011ebook}, as 0.1. 

        \textbf{Shallow Water Equations (SWE)}. The shallow water equations are derived by depth-integrating the incompressible Navier-Stokes equations and find applications in modeling atmospheric flows~\citep{vreugdenhil1994numerical}.
        We use the dataset from~\citet{gupta2023towards} for modeling the velocity and pressure fields for global atmospheric winds with a periodic boundary condition. 
        % The trajectories are generated using the numerical solver from~\citet{klower2milankl} on a $96\times 192$ spatial discretization. Each trajectory consists of $11$ time steps, with $48$ hours between each step. The data are split as 5,600/1,400/1,400, and $h=2$ historical time steps are used such that trajectories are unrolled for 9 steps during evaluation. 
    
    \subsection{Models}\label{sec:models}

% \todo[inline,size=\tiny]{Appendix refs; if we don't include the small data, change SineNet-K to SineNet-8}
        Here, we overview the models used in our experiments%al comparisons
, and provide further details in Appendix~\ref{appdx:implementation}.
        
        \textsc{\textbf{SineNet-8/16.}} SineNet with 8 or 16 waves and $64$ channels at the highest resolution. Multiplier $m_k=1.425/1.2435$ is used, resulting in channels along the downsampling path of each wave being arranged as $(64,91,129,185,263)/(64,79, 98, 123, 153)$.

        % \textsc{\textbf{SineNet-$K$-128.}} \textsc{SineNet-$K$} with $128$ channels at all resolutions, \textit{i.e.}, $m_K=1$. 

        \textsc{\textbf{F-FNO.}} Fourier Neural Operators~\citep{li2021fourier} process multi-scale features by performing convolutions in the frequency domain~\citep{gupta2023towards}, and were originally developed for PDE modeling. As one of the primary components of SineNet is depth, we compare to a state-of-the-art FNO variant, the Factorized FNO (\textsc{F-FNO})~\citep{tran2023factorized}, optimized specifically for depth. In each of the $24$ layers, the number of Fourier modes used is $32$ and the number of channels is $96$.
        
        \textsc{\textbf{Dil-ResNet.}} As opposed to downsampling and upsampling, the dilated ResNet (\textsc{Dil-ResNet}) proposed by~\citet{stachenfeld2022learned} is a neural PDE solver that processes multi-scale features by composing blocks of convolution layers with sequentially increasing dilation rates followed by sequentially decreasing dilation rates~\citep{zhang2023artificial}. 

        \textbf{\textsc{U-Net-128} and \textsc{U-Net-Mod}.} \textsc{U-Net-128} is equivalent to \textsc{SineNet-1} with the multiplier $m_K=2$ and $128$ channels at the highest resolution, with channels along the downsampling path arranged as $(128,256, 512,1024,2048)$.
        \textsc{U-Net-Mod} is a modern %ified 
        version of the
        % conventional
        U-Net architecture which parameterizes downsampling using strided convolutions and upsampling using transposed convolutions. Additionally, it doubles the number of convolutions and skip connections per resolution. 
        
        % \textsc{\textbf{U-Net-128.}} This architecture is equivalent to \textsc{SineNet-1} with the multiplier $m_K=2$ and $128$ channels at the highest resolution, with channels along the downsampling path arranged as $(128,256, 512,1024,2048)$.

        % \textsc{\textbf{U-Net-Mod.}} A modified version of the conventional U-Net architecture which parameterizes downsampling using strided convolutions and upsampling using transposed convolutions. Additionally, doubles the number of convolutions and skip connections per resolution.   

        For convolution layers in \textsc{SineNet%-8
        , Dil-ResNet, U-Net-128} and \textsc{U-Net-Mod}, zero padding is used on the INS dataset and circular padding is used on CNS and SWE.

    \subsection{Results}\label{sec:results}
    % \subsection{Results and ablation study}\label{sec:results}

        % \todo[inline,size=\tiny]{Add app ref}
        
        We present results for INS, CNS, and SWE in Table~\ref{tab:res}. On all datasets, \textsc{SineNet-8} has the lowest 1-step and rollout errors. Out of the remaining baselines, F-FNO has the strongest performance in terms of rollout error on CNS and SWE, while \textsc{Dil-ResNet} has the best baseline rollout error on INS. We visualize \textsc{SineNet-8} predictions on each dataset in Appendix~\ref{app:pred_vis}.

        \begin{table}[h]
            \caption{Summary of rollout test error and one-step test error. The best performance is shown in bold and the second best is underlined.}
            \vspace{-7pt}  % -7
            % \caption{Test errors. The best performance is shown in bold and the second best is underlined.}
            \begin{center}
            \label{tab:res}
                \begin{sc}
                    \resizebox{\textwidth}{!}{
    \begin{tabular}{lccccccc}
        \toprule
        & & \multicolumn{2}{c}{INS} & \multicolumn{2}{c}{CNS} & \multicolumn{2}{c}{SWE} \\%\cline{3-6}
        Method & \# Par. (M) & 1-Step ($\%$) & Rollout ($\%$) & 1-Step ($\%$) & Rollout ($\%$) & 1-Step ($\%$) & \textsc{Rollout ($\%$)} \\
        \midrule
        SineNet-8 & 35.5 & \underline{1.66} & \underline{19.25} & \underline{0.93} & \underline{2.10} & \underline{1.02} & \underline{1.78} \\
        SineNet-16 & 35.4 & \textbf{1.46} & \textbf{17.63} & \textbf{0.87} & \textbf{2.04} & \textbf{0.92} & \textbf{1.64}
        \\
        % SineNet-8-128 & 28.3 & \textbf{1.41} & \textbf{17.15} & \textbf{0.82} & \underline{2.15} & \textbf{0.93} & \textbf{1.66} \\
        F-FNO & 30.1 & 2.41 & 22.58 & 1.46 & 2.91 & 1.22 & 2.46 \\
        Dil-ResNet & 16.7 & 1.72 & \blue{19.29} & \blue{1.17} & \blue{3.76} & \blue{2.23} & \blue{4.12} \\
        U-Net-128 & 135.1 & 2.69 & 24.94 & 1.62 & 3.05 & 1.63 & 3.38 \\
        U-Net-Mod & 144.3 & 2.43 & 23.65 & 1.32 & 3.34 & 1.60 & 3.02 \\
        \bottomrule
    \end{tabular}

                    }
                \end{sc}
            \vspace{-11pt}  % -11
            \end{center}
        \end{table}

    \subsection{Ablation study}\label{sec:ablation}
            \begin{table}[h]
            \caption{Ablation study results. The best performance is shown in bold.}
            \vspace{-7pt}  % -7
            \begin{center}
            \label{tab:abl}
                \begin{sc}
                    \resizebox{\textwidth}{!}{
    \begin{tabular}{lccccccc}
        \toprule
        & & \multicolumn{2}{c}{INS} & \multicolumn{2}{c}{CNS} & \multicolumn{2}{c}{SWE} \\%\cline{3-6}
        Method & \# Par. (M)  & 1-Step ($\%$) & Rollout ($\%$) & 1-Step ($\%$) & Rollout ($\%$) & 1-Step ($\%$) & Rollout ($\%$)\\
        \midrule
        SineNet-8 & 35.5 & \textbf{1.66} & \textbf{19.25} & \textbf{0.93} & \textbf{2.10} & \textbf{1.02} & \textbf{1.78} \\
        SineNet8-Entangled & 32.8 & 1.69 & 19.46 & 1.01 & 2.39 & 1.14 & 1.97 \\
        Deeper U-Net-8 & 28.6 & 1.74 & 19.58 & 1.21 & 2.76 & 1.39 & 2.67 \\
        % \midrule
        % SineNet-8-128 & 28.3 & \textbf{1.41} & \textbf{17.15} & \textbf{0.82} & 2.15 & \textbf{0.93} & \textbf{1.66} \\
        % Deeper U-Net-8-128 & 20.1 & 1.46 & 17.22 & 0.84 & \textbf{1.92} & 1.33 & 2.51 \\
        \bottomrule
    \end{tabular}

                    }
                \end{sc}
                            \vspace{-10pt}  % -10
            \end{center}
        \end{table}    

\begin{figure}
    \centering
    \includegraphics[width=\textwidth]{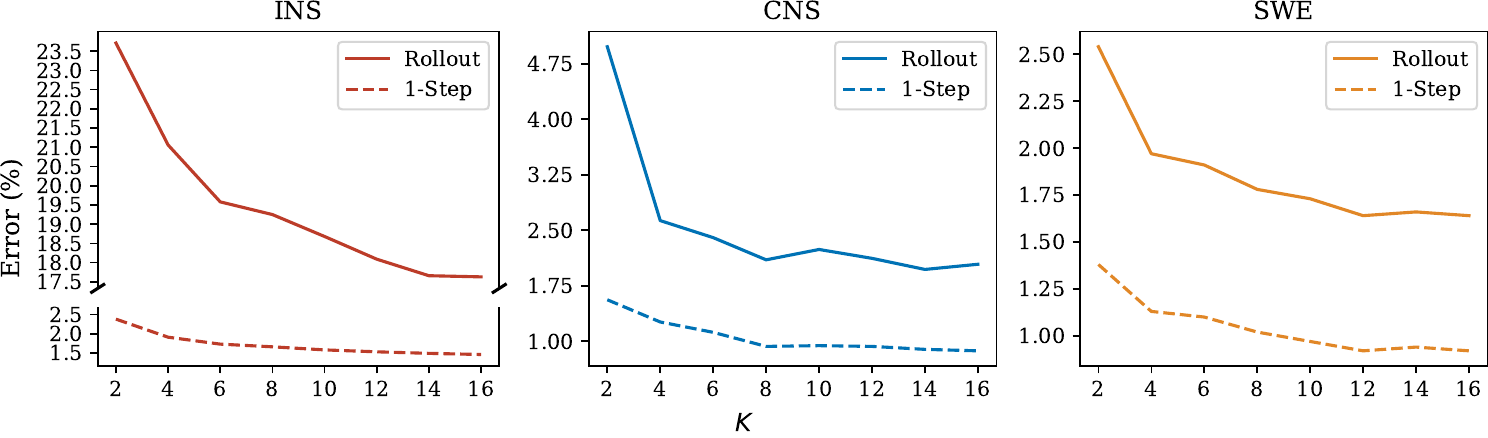}
    % \vspace{-20pt}
    \caption{Number of waves $K$ versus test error. Solid line is rollout error, dashed line is 1-step error. \blue{Numerical results are presented in Appendix~\ref{app:numres}}.}
    \vspace{-17pt}
    \label{fig:k_abl}
\end{figure}

        We construct several models to ablate various components of SineNet and present results for these models in Table~\ref{tab:abl}. 
        
        \textsc{\textbf{Deeper U-Net-8.}} Conventional U-Net, but increases the number of convolutions per resolution along the down and upsampling paths by a factor of $8$ relative to \textsc{SineNet-$8$}, resulting in an architecture with a similar number of layers to \textsc{SineNet-$8$}, as visualized in Figure~\ref{fig:models} in Appendix~\ref{appdx:implementation}. This serves to ablate SineNet's multi-stage processing strategy as a means of resolving the misalignment issue, and furthermore validate that observed improvements of SineNet are not a mere result of increased depth. In Table~\ref{tab:abl}, we see that on all datasets, \textsc{SineNet-8} outperforms \textsc{Deeper U-Net-8}. The largest improvements are on the SWE data, where the time interval between consecutive solutions is 48 hours. While the large timestep results in substantial misalignment in the \textsc{Deeper U-Net} skip connections, the multi-stage strategy of \textsc{SineNet} handles it effectively. 
        
        \textsc{\textbf{SineNet-8-Entangled.}} \textsc{SineNet-8}, but the parallel and sequential processing branches are entangled, \textit{i.e.}, the downsampling path is constructed as in Equation~\ref{eq:bres_down} as opposed to the disentangled downsampling in Equation~\ref{eq:disent}. In Table~\ref{tab:abl}, we observe that disentangling the parallel branch results in consistent performance gains.
        
        \textbf{\textsc{SineNet-$K$}.}
        \textsc{SineNet} with $K$ waves, and the channel multiplier $m_K$ chosen such that the number of parameters is roughly constant across all models, which we discuss further in Appendix~\ref{sec:channels}. In Figure~\ref{fig:k_abl}, we present results for $K=2,4,6,8\blue{,10,12,14,16}$ and find that on all $3$ datasets, errors monotonically improve with $K$, \blue{although improvements appear to plateau around $K=16$}. This result is consistent with our assumption of the latent evolution of features, as a greater number of waves leads to a smaller amount of evolution managed by each wave, thereby leading to improved modeling accuracy through reduced misalignment. In Appendix~\ref{app:ts}, we present the inference time and space requirements for \textsc{SineNet}-$K$.

\begin{wraptable}{r}{0.45\linewidth}
    \vspace{-0.45cm} % 0.4
    \centering
    \caption{Comparison between zero and circular padding on SWE with \textsc{SineNet-8}.}
    % \vspace{-9.5pt}  % 0
    \begin{sc}
     \resizebox{0.45\textwidth}{!}{
    \begin{tabular}{cccc}
        \toprule
        \multicolumn{2}{c}{Zero padding} & \multicolumn{2}{c}{Circular padding} \\
          \textsc{1-Step ($\%$)} & \textsc{Rollout ($\%$)} & \textsc{1-Step ($\%$)} & \textsc{Rollout ($\%$)} \\\midrule
         1.50 & 4.19 & \textbf{1.02} & \textbf{1.78} \\
        \bottomrule
    \end{tabular}}
    \label{tab:padding}
    \end{sc}
    % \vspace{-.5cm}
    % \vspace{-20pt}
\end{wraptable}
\textbf{Effect of circular padding.} In Table~\ref{tab:padding}, we replace circular padding with zero padding in \textsc{SineNet-8} on SWE. The periodic boundaries result in boundary points opposite each other being spatially close, however, without appropriate padding to encode this into feature maps, the ability to model advection across boundaries is severely limited.

\colorblue

% \section{Discussion \blue{and Conclusion}}
\section{Discussion}

We discuss potential limitations in Appendix~\ref{app:limitation}, including out-of-distribution (OOD) generalization, applicability to PDEs with less temporal evolution, computational cost, and handling irregular spatiotemporal grids. 
Additionally, while SineNet is a CNN and therefore cannot directly generalize to discretizations differing from that in the training set, inference at increased spatial resolution can be achieved through dilated network operations or interpolations without any re-training, which we discuss and experiment with further in Appendix~\ref{app:sr}.
\color{black}

\section{Conclusion}

% \section{Discussion and conclusion}
% \todo[inline, size=\tiny]{check note about ablation}

% \blue{
% \textbf{Limitations.} We discuss the potential limitations of our method in Appendix.
% }
In this work, we have presented SineNet, a neural PDE solver designed to evolve temporal dynamics arising in time-dependent PDEs. We have identified and addressed a key challenge in temporal dynamics modeling, the misalignment issue, which results in a performance drop for conventional U-Nets. By reframing the U-Net architecture into multiple waves, SineNet mitigates this misalignment, leading to consistent improvements in solution quality over a range of challenging PDEs. Further, we analyze the role of skip connections in enabling both parallel and sequential processing of multi-scale information. Additionally, we demonstrate that increasing the number of waves, while keeping the number of parameters constant, consistently improves the performance of SineNet. Our empirical evaluation across multiple challenging PDE datasets highlights the effectiveness of SineNet and its superiority over existing baselines.

\section*{Acknowledgments}
This work was supported in part by National Science Foundation grants IIS-2243850 and IIS-2006861.

\bibliography{reference}
\bibliographystyle{iclr2024_conference}

\newpage

\appendix
\section*{Appendix}
% \todo[inline, size=\tiny]{We need a time/space analysis; add some details for each of the model architectures

% Need to mention normalization
% }

\section{Misalignment analysis}

\subsection{Slowing down evolution in latent space}\label{sec:slow}

% \todo[inline, size=\tiny, color=pink]{Maybe it's more suitable to put this section somewhere in experiments instead of method?}

\begin{wrapfigure}[22]{r}{0.45\linewidth}
% \begin{figure}[h]
    \vspace{-10pt}
    \centering
    \includegraphics[width=\linewidth]{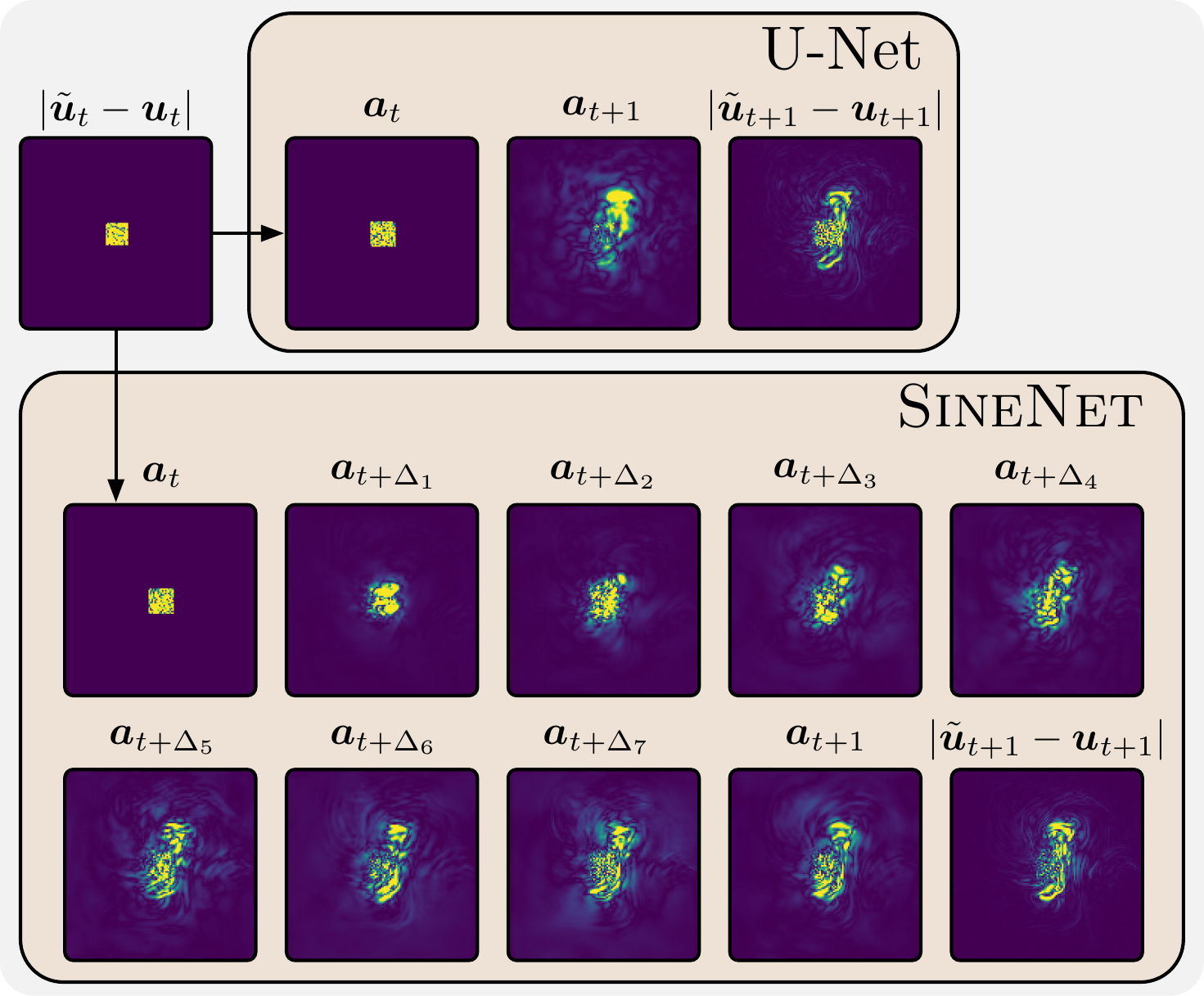}
    \caption{Feature map responses to noise injection for trained \textsc{U-Net-128} and \textsc{SineNet-8}. As opposed to the \textsc{U-Net}, in \textsc{SineNet}, the perturbation influence propagates gradually from the first to the last feature map, demonstrating the reduced latent evolution managed by each wave.}
    \label{fig:feature_difference}
% \end{figure}
\end{wrapfigure}

We now conduct an empirical analysis to validate our hypothesis concerning the manageable inconsistency due to reduced misalignment within each wave. Specifically, we demonstrate the progressive evolution of the system's state over the waves by injecting a perturbation into the model input and monitoring the ensuing response in the latent maps at each wave. The response to this perturbation serves as an indicator of the influence of the perturbation at each time step, thereby enabling us to track the evolution of the system. To quantify this response, we first carry out a forward pass of the input $\bm u_t$ through our trained model and capture the $\ell$-th feature map, denoted as $\bm x_{t+\Delta \ell}$. We then introduce a localized perturbation to $\bm u_t$ in the form of random noise concentrated in a small central region and re-record the $\ell$-th feature map, denoted as $\tilde{\bm x}_{t+\Delta \ell}$. The absolute difference $\bm a_{t+\Delta \ell}\coloneqq|\tilde{\bm x}_{t+\Delta \ell} - \bm x_{t+\Delta \ell}|$ is computed and visualized to represent the response.
\blue{We add such noise to samples from the test set as it allows us to see the feature propagation from an excitation with a small spatial extent. Otherwise, if we only provide the perturbation as input to the network, it will become an out-of-distribution problem for the network since there are no similar samples in the training set. However, if we inject a small amount of noise to a sample in the test set, the resulting input remains close to the original test sample, and therefore is still in-distribution and can be used to analyze the behavior of the trained model.}
% \todo[inline,size=\tiny]{Figure 3 shows artifacts from transposed conv; we use interpolation now. Fig 3 uses $x$ instead of $a$}

Figure~\ref{fig:feature_difference} showcases the evolution of this response across the latent maps $\bm a_{t+\Delta_0}, \ldots, \bm a_{t+\Delta_K}$ in a well-trained \textsc{SineNet-8} model. For comparison, we also include the response from a well-trained \textsc{U-Net-128}, using the output of the first projection layer and the output of the last upsampling block. In both the \textsc{U-Net} and \textsc{SineNet} models, the perturbation influence expands spatially from the input to the output. However, for \textsc{SineNet-8}, the perturbation influence propagates incrementally from the first to the last feature map, indicating that the spatial evolution within each wave is reduced. It's important to note that this progressive propagation is not merely a consequence of limited receptive fields, as the receptive field of each wave is substantially %ignificantly 
larger than the area influenced by the perturbation. Furthermore, the receptive field for each wave of \textsc{SineNet} is equal to that of the \textsc{U-Net}.

% To summarize, SineNet defines a learned operator in the spatial domain and is ideally suited for learning temporal dynamics.
% % % % % % % % % % 
\subsection{Feature map visualization}\label{app:fmvis}
% \todo[inline,size=\tiny]{Update}
We visualize feature maps for each wave of \textsc{SineNet-8} in Figure~\ref{fig:feature_vis} for a randomly selected example from the INS data. Specifically, for each wave, we show the highest-resolution skip connection from the downsampling path alongside the result of the upsampling path to which the skip-connected feature map is concatenated, as due to latent evolution, these two feature maps will have the greatest misalignment out of all of the skip connections in a given wave. 
% To get an intuitive understanding for the evolution of feature maps, we visualize and compare the feature maps from different layers of trained models (Figure~\ref{fig: misalignment},~\ref{fig: sinenet},~\ref{fig:feature_difference} in main text and Figure~\ref{fig:feature_vis} in the Appendix). 
Visualization is done by averaging over the channel dimension of each feature map. 
% We find the mean feature maps exhibit an interesting visual resemblance to fluid fields. This could be interpreted by the localization of feature activations. For instance, regions with rich texture in the fluid field often result in rich texture in the feature map. Feature maps could have extreme values in some small regions, and thus, to 
To increase the overall contrast, we clip the 1\textsuperscript{st} and 99\textsuperscript{th} quantiles.
% We clip at (0.001, 0.999) quantiles for Figure~\ref{fig:feature_difference}, (0, 0.997) quantiles for Figure~\ref{fig:feature_difference} and (0.01, 0.98) quantiles for Figure~\ref{fig: sinenet} and Figure~\ref{fig:feature_vis}.
% Figure~\ref{fig:feature_vis} visualizes the feature maps related to the skip connections at the top of wave in SineNet-6. 
As can be seen relative to the feature maps from the U-Net in Figure~\ref{fig: misalignment}, the misalignment in skip connections is substantially
% ignificantly 
mitigated thanks to the multi-stage feature evolution in SineNet.

\afterpage{
\begin{figure}[p!]
    \centering
    \includegraphics[height=0.9\textheight]{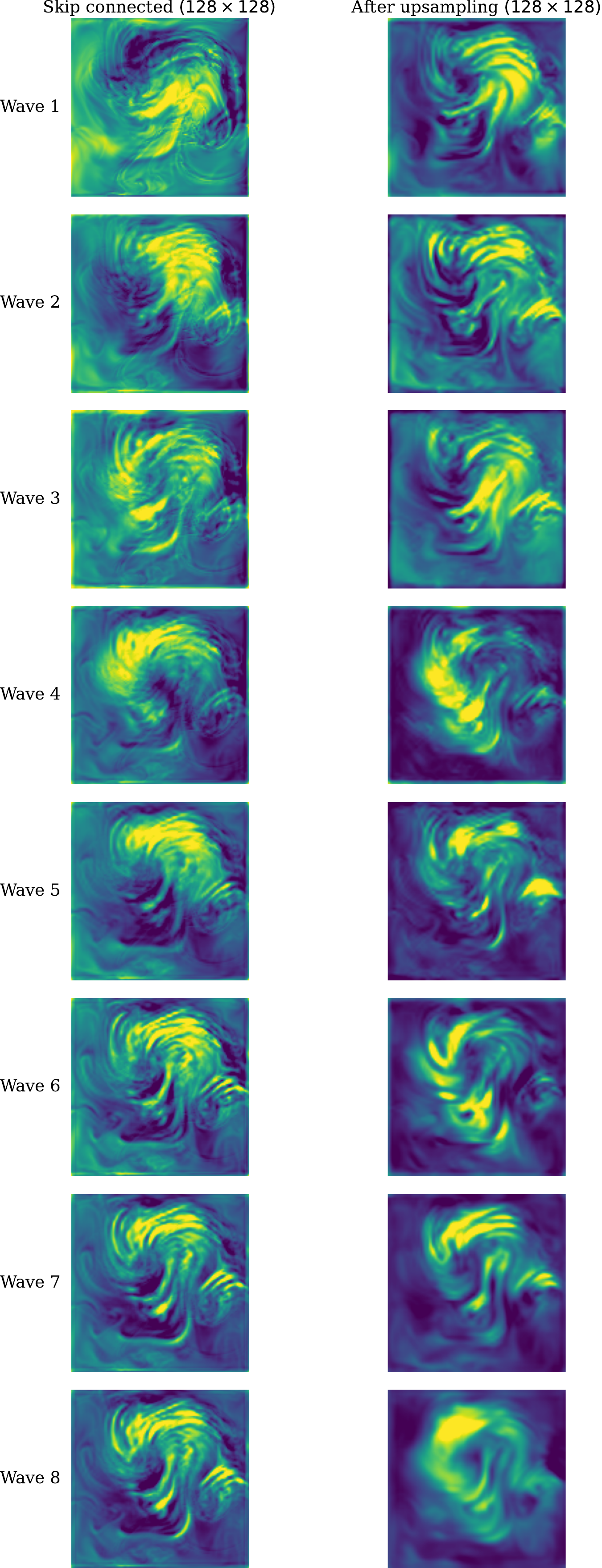}
    \caption{Visualization of \textsc{SineNet-8} feature maps averaged over the channel dimension for INS data.  
    Skip connections from the highest resolution in the downsampling path of each wave are visualized on the \textbf{left}, which are concatenated with the upsampled feature maps visualized on the \textbf{right}. 
    % from the   with the at the last upsampling block of each wave are visualized on the left, where feature maps (\textbf{left}) are first upsampled with transposed convolution (\textbf{middle}) and are consequently concatenated with the input of the current wave (\textbf{right}).   The spatial resolutions of feature maps is indicated in parentheses. $L$ denotes the number of the wave (U-Net block) at which the visualized feature map appears. Feature maps are averaged over the channel dimension for visualization. 
    Compared to U-Net (Figure~\ref{fig: misalignment} in main text), the degree of misalignment in skip connections is far less severe, as feature maps progressively evolve more gradually across waves.}
    \label{fig:feature_vis}
\end{figure}
\clearpage
}

\blue{
\section{Dataset details}}

\subsection{The Navier-Stokes equation}\label{}
The Navier-Stokes equations \citep{constantin2020navier} are the fundamental model which encodes the conservation of mass and momentum in fluid flows in manifold applications ranging from weather research to aerospace engineering and stellar magneto-hydrodynamics. It comprises a system of non-linear partial differential equations for the density $\rho$, velocity field $\bm u$, and pressure $p$ of a fluid (compressible case) or velocity $\bm u$ and pressure $p$ (incompressible case). The Navier-Stokes equations are the dominant model of fluid dynamics from which other equations such as the Euler equation or the shallow water equations are derived in limiting regimes (zero viscosity, shallow water). Due to its foundational role, it is the subject of an unsolved Millennium Problem of the Clay Mathematics Institute and has attracted attention from mathematical analysts as well as applied scientists. Challenges abound, especially for fast-moving turbulent flows, and despite recent breakthroughs \citep{albritton2022non}, even the existence and uniqueness of solutions is not fully understood. 
% Given its fundamental importance, challenging nature, and wide range of applications, we select this PDE for the focus of our work.

\blue{\subsection{Detailed setup}}\label{sc:dataset_details}

 \textbf{Incompressible Navier-Stokes (INS)}. 
 % The incompressible Navier-Stokes equations model the flow of a fluid wherein the density is assumed to be independent of the pressure, but may not be constant due to properties of the fluid such as salinity or temperature~\citep{vreugdenhil1994numerical}. The equations are given by 
 %        \begin{equation}\label{eq:ns}
 %            \frac{\partial \bm{v}}{\partial t} = -\bm{v}\cdot \nabla \bm{v} + \mu \nabla^2 \bm{v} - \nabla p + \bm{f},\ \nabla \cdot \bm{v} = 0,
 %        \end{equation}
 %        where $\bm{v}$ is velocity, $p$ is internal pressure, and $\bm{f}$ is an external force. 
        % We use the dataset from~\citet{gupta2023towards}, simulated with a numerical solver from the $\Phi_{\mbox{\tiny{Flow}}}$ package~\citep{holl2020phiflow}. 
        Each trajectory is 14 time steps spaced 1.5 seconds apart and models the evolution of a vector velocity field with a Dirichlet boundary condition and a scalar field with a Neumann boundary conditon representing the concentration of particles advected by the velocity field. Trajectories are spatially discretized onto a $128\times 128$ grid. The data follow a train/valid/test split of 5,200/1,300/1,300, and the length of the time history $h$ is 4 steps such that trajectories are unrolled for 10 steps during evaluation. The viscosity $\mu$ controlling the level of turbulence in the flow is $0.01$.

        % Anderson: The Mach number is the ratio of the flow velocity to the speed of sound
        % Anderson: A flow in which the density ρ is constant is called incompressible. In contrast, a flow where the density is variable is called compressible.
    \textbf{Compressible Navier-Stokes (CNS)}. 
    % Compressibility is generally a consideration most relevant for fast-moving fluids~\citep{anderson2011ebook}.
    % We generate our CNS dataset using the numerical solver from~\citet{takamoto2022pdebench}. 
    Each trajectory has 21 time steps and consists of a scalar pressure field, a scalar density field, and a vector velocity field, each with periodic boundary conditions. Trajectories are generated on a $512\times 512$ spatial grid and downsampled to $128\times 128$. The dataset is split as 5,400/1,300/1,300 and, following~\citet{takamoto2022pdebench}, models use time history $h=10$ such that trajectories are unrolled for $11$ steps during evaluation. 
    % The dynamics in this data are more turbulent than those in the INS data, with the viscosity as $1\times10^{-8}$ and the initial Mach number, which quantifies the ratio of the flow velocity to the speed of sound in the fluid~\citep{anderson2011ebook}, as 0.1. 

    \textbf{Shallow Water Equations (SWE)}. 
    % The shallow water equations are derived by depth-integrating the incompressible Navier-Stokes equations and find applications in modeling atmospheric flows~\citep{vreugdenhil1994numerical}. 
    % We use the dataset from~\citet{gupta2023towards} for modeling the velocity and pressure fields for global atmospheric winds with a periodic boundary condition. 
    The trajectories are generated using the numerical solver from~\citet{klower2milankl} on a $96\times 192$ spatial discretization. Each trajectory consists of $11$ time steps, with $48$ hours between each step. The data are split as 5,600/1,400/1,400, and $h=2$ historical time steps are used such that trajectories are unrolled for 9 steps during evaluation.

\section{Time and space complexity}\label{app:ts}

In Table~\ref{tab:ts}, we analyze the inference time\blue{, training time,} and GPU footprint for all models on a batch of size 32 randomly selected from the INS dataset. The times presented here are an average over 1,000 batches \blue{on a single 80GB A100 GPU}.

\begin{table}[h]
    \centering
    % \vspace{0.1in}
    \caption{Analysis of inference \blue{and training} time\blue{, as well as} memory required for all models \blue{on a batch of size 32 randomly selected from the INS dataset}. \blue{A batch size of 16 is used for \textsc{SineNet-Neural-ODE} in the Forward+Backward complexity test as it cannot fit into the GPU memory with batch size 32.}}\label{tab:ts}
    % \vspace{-5pt}
    % \small
    \begin{sc}
    \resizebox{\columnwidth}{!}{
     \begin{tabular}{lcccccc}
        \toprule
        % Method & Inference Time (s) & Forward Memory (GB) \\
        & & \multicolumn{2}{c}{\blue{Forward}} & \multicolumn{2}{c}{\blue{Forward+Backward}} \\
        Method & \blue{\# Par. (M)} & \blue{Time (s)} & \blue{Memory (GB)} & \blue{Time (s)} & \blue{Memory (GB)}\\
        \midrule
        SineNet-2 & \blue{35.5} & $0.122$ & $4.19$ &\blue{0.398} & \blue{10.63}\\
        SineNet-4 & \blue{35.5} & $0.180$ & $3.86$ &\blue{0.596} & \blue{15.11}\\
        SineNet-6 & \blue{35.5} & $0.225$ & $3.74$ &\blue{0.729} & \blue{19.18}\\
        SineNet-8 & \blue{35.5} & $0.286$ & $3.54$ &\blue{0.917} & \blue{23.31}\\
        \blue{SineNet-10} & \blue{35.5} & \blue{0.328} & \blue{3.52} &\blue{1.043} & \blue{27.21} \\
        \blue{SineNet-12} & \blue{35.5} & \blue{0.368} & \blue{3.87} & \blue{1.157} &  \blue{30.90}\\
        \blue{SineNet-14} & \blue{35.4} &\blue{0.410} &\blue{3.70} & \blue{1.286} & \blue{34.61} \\
        \blue{SineNet-16} & \blue{35.5} & \blue{0.447} & \blue{3.65} & \blue{1.400} & \blue{38.21} \\
        \blue{SineNet-Neural-ODE} & \blue{12.7} & \blue{1.822} & \blue{8.60} &\blue{2.977\textsuperscript{*}} & \blue{49.53\textsuperscript{*}}\\
        F-FNO  & \blue{30.1} & $0.399$ & $3.91$ &\blue{1.175} & \blue{41.05} \\
        Dil-ResNet-128 & \blue{4.2} & $0.182$ & $3.83$ &\blue{0.387} & \blue{28.77} \\
        \blue{Dil-ResNet-256} & \blue{16.7} & \blue{0.469} & \blue{8.37} &\blue{1.044} & \blue{56.93}\\
        U-Net-128 & \blue{135.1} & $0.169$ & $7.58$ &\blue{0.532} & \blue{13.24}\\
        U-Net-Mod & \blue{144.3} & $0.097$ & $4.71$ &\blue{0.236} & \blue{14.12} \\
        \bottomrule
    \end{tabular}
    }
    \end{sc}
    % \vspace{-5pt}
    \label{tab:lr}
\end{table}

\section{Implementation details}\label{appdx:implementation}

\begin{figure}[t]
    \centering
    \hspace*{\fill}%
    \subcaptionbox{\sc{SineNet}\label{fig:models_sinenet}}{\includegraphics[height=.9in]{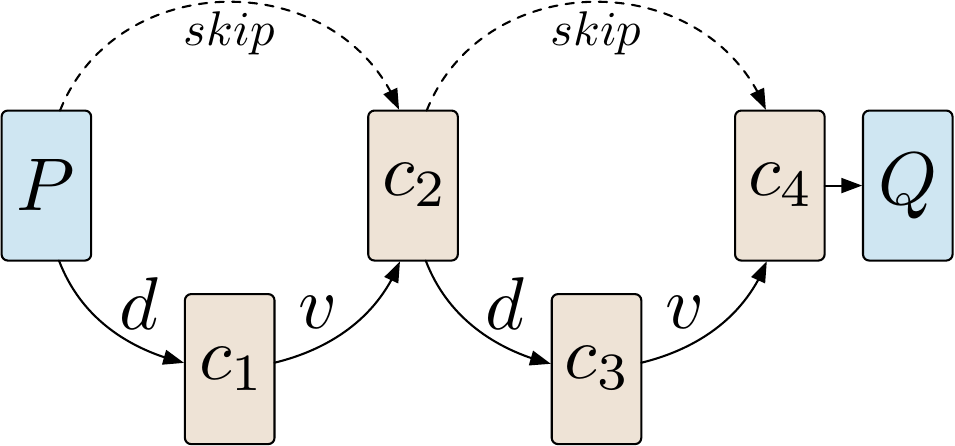}}\hfill%
    \subcaptionbox{\sc{U-Net}\label{fig:models_unet}}{\includegraphics[height=.9in]{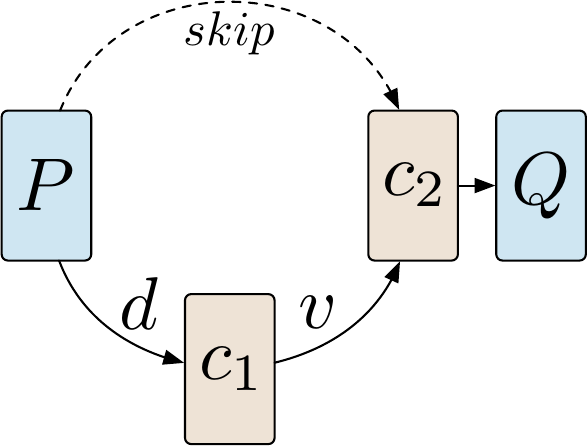}}\hfill%
    \subcaptionbox{\sc{Deeper U-Net}\label{fig:models_deeper}}{\includegraphics[height=.9in]{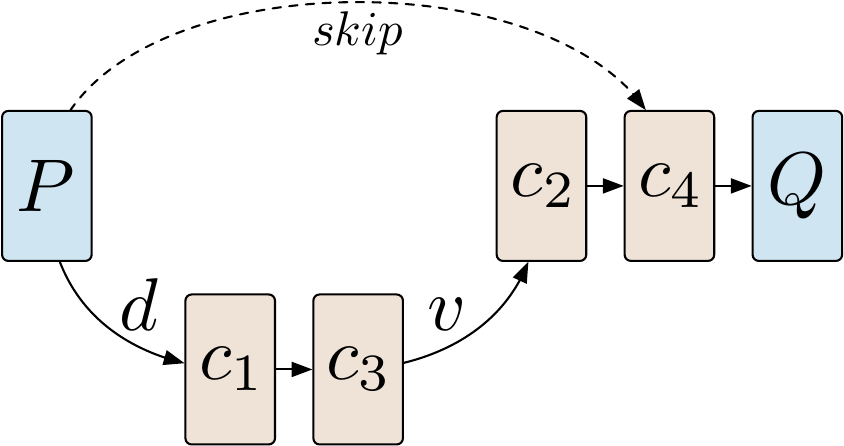}}%
    \hspace*{\fill}%
    \caption{Comparison of \textsc{SineNet}, \textsc{U-Net} and \textsc{Deeper U-Net} architectures.}
    \label{fig:models}
\end{figure}

\subsection{Training}\label{app:train}

Our code is implemented in PyTorch~\citep{paszke2019pytorch}. Models are trained and evaluated on $2$ NVIDIA A100 80GB GPUs. All models are optimized for 50 epochs with batch size 32 and the model with the best validation rollout results is used for testing. Following~\citet{gupta2023towards}, each epoch consists of $T$ iterations over the data. For each trajectory, a start time $t$ is randomly sampled from $t=h,\ldots,T-1$. Performing $T$ cycles per epoch ensures that each possible 1-step input-target pair $\left(\{\bm{u}_{t-h+1}, \ldots, \bm{u}_{t}\}, \bm{u}_{t+1}\right)$ is sampled more than once in expectation for each trajectory.  
% We use a default batch size of 32. 
% For all models performing convolutions in the spatial domain, we include circular padding for the datasets with periodic boundaries. 
We calculate statistics from the training data along the field dimension and normalize model inputs and targets to have $0$ mean and unit variance. During rollouts, we apply the inverse normalization. 

All models are optimized with the AdamW optimizer~\citep{kingma2014adam, loshchilovdecoupled}, using an initial learning rate of $\eta_{\text{init}}=2\times10^{-4}$, except for \blue{the \textsc{F-FNO} on SWE and CNS}, where we found a larger learning rate to improve performance \blue{(see Table~\ref{tab:hp})}. The learning rate was warmed up linearly for 5 epochs from $\eta_{\min}=1\times10^{-7}$ to $\eta_{\text{init}}$ before being decayed for the remaining 45 epochs using a cosine scheduler.

\subsection{Loss}\label{app:loss}

We use the Scaled-$L_2$ loss for both training and evaluation. We observe that the magnitude can vary significantly across different fields. Therefore, we compute the Scaled-$L_2$ loss separately for different fields to account for differences in magnitude and then calculate the average across all fields. Given a prediction $\hat{\bm{u}}_t$ and a target $\bm{u}_t$ at time step $t$, composed of $M$ fields, the 1-step loss is computed as
\begin{equation}\label{eq:loss_step}
    \mathcal{L}^{\text{1-step}}(\hat{\bm{u}}_t, \bm{u}_t) = \frac{1}{M}\sum_{k=1}^{M}\frac{\lVert\hat{\bm{u}}^k_t - \bm{u}^k_t\rVert_2}{\lVert \bm{u}^k_t\rVert_2},
\end{equation}
where $\bm{u}^k$ denotes the $k$-th field of $\bm{u}$ and the norm is taken over all spatial dimensions. Note that we consider each scalar component as a separate field. For example, the velocities along $x$ direction and $y$ direction are considered as two different fields and are normalized independently.
For validation and test, the rollout loss is computed as the average 1-step loss over rollout time steps
\begin{equation}\label{eq:loss_rollout}
    \mathcal{L}^{\text{rollout}} = \frac{1}{T-h} \sum_{t=h+1}^{T}\mathcal{L}^{\text{1-step}}(\hat{\bm{u}}_t, \bm{u}_t),
\end{equation}
where $T$ is the number of total time steps and $h$ is the number of conditioning historical time steps.

\subsection{Selection of multiplier hyperparameter}\label{sec:channels}

\begin{table}[h]\label{tab:multiplier}
    \centering
    % \vspace{0.1in}
    \caption{Multiplier hyperparameter and number of channels in each wave for \textsc{SineNet-$K$} across various choices of $K$. The multiplier determines how the number of channels is upscaled along the downsampling path and downscaled along the upsampling path. We adjust this multiplier to manage the number of parameters and computational cost of SineNet. We also list the number of channels produced by each of the 4 blocks from the beginning to the end of the downsampling path. The input number of channels to each wave in all models is $64$.}
    % \vspace{-5pt}
    % \small
    \begin{sc}
    % \resizebox{0.85\columnwidth}{!}{
     \begin{tabular}{lccc}
        \toprule
        Method & \# Par. (M) & Multiplier & \# of Channels\\
        \midrule
        SineNet-2 & $35.5$ & $1.8075$ & $\left(115,209,377,683\right)$\\
        SineNet-4 & $35.5$ & $1.6110$ & $\left(103,166,267,431\right)$\\
        SineNet-6 & $35.5$ & $1.5000$ &  $\left(96,144,216,324\right)$\\
        SineNet-8 & $35.5$ & $1.4250$ & $\left(91,129, 185, 263\right)$\\   
        \blue{SineNet-10} & \blue{$35.5$} & \blue{$1.3660$} & \blue{$\left(87,119,163,222\right)$}\\ 
        \blue{SineNet-12} & \blue{$35.5$} & \blue{$1.3190$} & \blue{$\left(84,111,146,193\right)$}\\ 
        \blue{SineNet-14} & \blue{$35.4$} & \blue{$1.2790$} & \blue{$\left(81,104,133,171\right)$}\\ 
        \blue{SineNet-16} & \blue{$35.5$} & \blue{$1.2435$} & \blue{$\left(79,98,123,153\right)$}\\ 
        \bottomrule
    \end{tabular}
    % }
    \end{sc}
    % \vspace{-5pt}
    \label{tab:sinenet_mult}
\end{table}

We now discuss the choice of the number of channels in the feature maps along the down and upsampling paths. The number of channels in the feature map output by the $\ell$-th downsampling and upsampling blocks, respectively, is given by
\begin{equation}
    z_\ell=\left\lfloor m_K^{\ell}z_0\right\rfloor,\qquad z_\ell=\left\lfloor m_K^{4-\ell}z_0\right\rfloor,\ell=1,\ldots,L
\end{equation}
where $z_0$ is the number of channels following the projection by the encoder $P$ and there are $L$ downsampling and upsampling blocks. While conventional U-Nets use multiplier $m_K=2$, we manage the number of parameters and complexity of our \textsc{SineNet-$K$} architectures by selecting $m_K$ such that the number of parameters is roughly constant in $K$. In Table~\ref{tab:sinenet_mult}, we present the multiplier $m_K$ and number of channels along the downsampling path for the \textsc{SineNet-$K$} architecture with varying number of waves $K$. All architectures presented here use $z_0=64$.

\subsection{Baseline Tuning}

    Here we present hyperparameter tuning results for baseline methods on the SWE dataset. As discussed in Appendix~\ref{app:train}, we found that for \textsc{F-FNO}, using a larger learning rate improved results. Additionally, \citet{tran2023factorized} found that in their setting, sharing weights between Fourier convolution layers improved F-FNO performance. However, we find not sharing to give better results. \blue{We additionally found that increasing the number of channels in \textsc{Dil-ResNet} improved performance}. We present validation errors for these experiments in Table~\ref{tab:hp}.

\begin{table}[h]
    \centering
    % \vspace{0.1in}
    \caption{\blue{Best validation results for different baseline hyperparameter settings on SWE. We evaluate \textsc{F-FNO} with different learning rates, as well as with and without weight sharing. We additionally evaluate the effect of number of channels on \textsc{Dil-ResNet}.}}\label{tab:hp}
    % \vspace{-5pt}
    % \small
    % \resizebox{0.85\columnwidth}{!}{
     \begin{tabular}{lcccc}
        \toprule
            % \begin{sc}
        \textsc{Method} & \textsc{\# Par. (M)} &   \textsc{1-Step Valid (\%)} & \textsc{Rollout Valid (\%)}\\
        % \end{sc}
        \midrule
        \textsc{FFNO}, $\eta_\text{init}=1\times10^{-3}$ & 30.1 & $\mathbf{1.21}$ & $\mathbf{2.44}$ \\
        \textsc{FFNO}, $\eta_\text{init}=2\times10^{-4}$ & 30.1 & $1.88$ & $3.72$ \\
        \textsc{FFNO-shared}, $\eta_\text{init}=1\times10^{-3}$ & 3.0 & $1.77$ & $3.44$ \\
        \midrule
        \textsc{Dil-ResNet}-128 & 4.2 & $3.40$ & $6.60$ \\
        \blue{\textsc{Dil-ResNet}-256} & \blue{$16.7$} & \blue{$\mathbf{2.21}$} & \blue{$\mathbf{4.09}$} \\
        \bottomrule
    \end{tabular}
    % }
    % \vspace{-5pt}
\end{table}

\colorblue
\bluecaption

\section{Discussion on limitations}\label{app:limitation}

\textbf{Out-of-distribution (OOD) generalization.}
% Classical solvers can solve any new initial condition. % Surrogate model can only generalize well when training distribution are similar
Unlike classical solvers which are based on intrinsic mathematical model of PDEs and can be applied to almost any initial and boundary conditions, data-driven surrogate models are only guaranteed to generalize well to scenarios which are close to the training distribution.
For example, performance drops substantially when simulating the same dynamics but on a larger domain than observed during training~\citep{stachenfeld2022learned}.
% Hybrid model could mitigate this. But it would be interesting to see how to improve OOD for surrogate models.
It is possible to improve OOD robustness by combining surrogate models with classical solvers~\citep{kochkov2021machine}.
However, pure surrogate models remain attractive due to flexibility in model design and fast inference.
It remains to be seen how training methods can be adapted to improve OOD robustness, with initial works in this direction focusing on meta-learning approaches~\citep{wang2022metalearning,mouli2023metaphysica,kirchmeyer2022generalizing}. In Appendix~\ref{app:cns120}, we found noise injection during training~\citep{sanchez2020learning,stachenfeld2022learned} improved robustness to difficult intial conditions and generalization to rollouts $10\times$ longer than training rollouts.

% \textbf{PDEs with less spatial evolution patterns.}
\textbf{PDEs with reduced temporal evolution.}
SineNet was developed specifically for time-evolving PDEs wherein the input fields are spatially misaligned with respect to the target fields. Therefore, further experimentation is needed to determine the benefits of SineNet in learning dynamics where time evolution does not play as large of a role, such as the steady-state Darcy flow equations considered by~\cite{li2021fourier}.

\textbf{Computational cost.} As shown in Appendix~\ref{app:ts}, using more waves increases the inference time and training memory due to the added depth. 
% Although the inference memory remains almost constant with more waves, the training memory will noticeably increase due to the added depth. 
Nevertheless, as we did in the experiments, channel multipliers can be adjusted to control the computational cost.

\textbf{Irregular spatial grids and time intervals.} For modeling dynamics on irregular discretizations, graph neural networks have been used ~\citep{brandstetter2022message,pfaff2021learning,wu2022learning}. As with all CNNs, SineNet can only handle inputs on rectangular grids with uniformly spaced mesh points. However, it is possible to extend SineNet to irregular meshes with graph neural networks~\citep{gao2019graph,li2020multipole}, where the key idea of combining multi-resolution processing and multi-stage processing remains valid. Furthermore, because SineNet is trained for autoregressive prediction, it can only advance time by a fixed-size timestep $\Delta_t$, and thus, cannot predict the solution at time points that are not multiples of $\Delta_t$. However, as we show in Appendix~\ref{app:conditional}, SineNet can be trained to generalize over $\Delta_t$.

\textbf{Mixed and non-null boundaries.} As discussed in Section~\ref{sec:bound}, the Dirichlet boundary conditions we consider for the particle concentration field can be effectively encoded via zero padding. Although the Neumann boundary conditions on the velocity field would ideally be encoded in feature maps using reflection padding to represent the null spatial derivative, encoding both the Neumann boundary condition and the Dirichlet boundary condition is not straightforward. This is because it is unclear which feature maps correspond to the particle concentration, and which correspond to the velocity field. To further complicate matters, feature maps earlier in the architecture may correspond to both. Although here we choose to only use zero-padding for simplicity, future work should explore principled approaches to encoding mixed boundary conditions for convolutional architectures, with \cite{horie2022physics} having initiated this line of work for graph neural networks.

Furthermore, while we consider null Dirichlet and null Neumann boundaries here, non-null conditions are also of interest. For non-null Dirichlet conditions, the value of the field on the boundary is known and can be encoded simply by padding with the known value. In the case of the non-null Neumann condition, one approach could be to pad with a finite difference-type approximation using the feature map's boundary value and the boundary derivative given by the condition. Further work is needed to validate the effectiveness of both approaches. 

\section{Extended results}

\subsection{Conditional INS}\label{app:conditional}

\begin{table}[h]
    \centering
    % \vspace{0.1in}
    \caption{Summary of rollout test error and one-step test error for various time step sizes on Conditional INS. }\label{tab:cond}
    % \vspace{-5pt}
    % \small
    \begin{sc}
    \resizebox{\columnwidth}{!}{
     \begin{tabular}{lcc|cccccc|c}
        \toprule
        & & & \multicolumn{5}{c}{1-Step (\%)} & \\
        Method & \# Par. (M) & & $\Delta_t=0.375$ & $\Delta_t=0.75$ & $\Delta_t=1.5$ & $\Delta_t=3.0$ & $\Delta_t=6.0$  & & Rollout (\%)\\
        \midrule
        SineNet-8-Add & 38.0 && 2.29 & \underline{2.82} & \underline{3.82} & \underline{6.84} & \underline{19.79} && 5.27 \\
        SineNet-8-AdaGN & 40.3 && \textbf{2.05} & \textbf{2.59} & \textbf{3.63} & \textbf{6.64} & \textbf{19.21} && \textbf{4.60} \\
        U-Net-Mod-Add & 146.6 && 2.46 & 3.23 & 4.94 & 9.50 & 24.69 && 5.85 \\
        U-Net-Mod-AdaGN & 148.8 && \underline{2.16} & 2.91 & 4.46 & 8.86 & 24.94 && \underline{5.00} \\
        % Dil-ResNet-Add & 25.6 && 4.79 & 5.81 & 8.05 & 13.65 & 29.51 && 11.71 \\
        \bottomrule
    \end{tabular}
    }
    \end{sc}
    % \vspace{-5pt}
\end{table}

    In the conditional task considered by~\cite{gupta2023towards}, models are trained to generalize over variable-sized time steps and variable forcing terms. Specifically, in Equation~\ref{eq:ns}, $\bm f$ is an external force acting along the $x$ and $y$ axes as $\bm f =(0,f)^\top\in\mathbb R^2$. In the conditional task, the $y$-component of the forcing term varies between trajectories as $f\in[0.2,0.5]$. Furthermore, the time step size is reduced by a factor of 4 relative to the INS data from 1.5 seconds to 0.375 seconds such that the number of time steps increases from 14 to 56. The one-step training objective then becomes the prediction of $\bm u_{t+\Delta_t}$ given $\bm u_t$, where the time step $\Delta_t$ takes values in $\{0.375k:k=1,\ldots,55\}$. 
    
    To model the variable forcing term and time step size, network inputs are $(\bm u_t,\kappa)$, where $\kappa$ is the time-step and buoyancy $\kappa\coloneqq(f,\Delta_t)$. \cite{gupta2023towards} condition models on $\kappa$ by learning an embedding $\epsilon$ of $\kappa$ obtained by applying sinusoidal embeddings~\citep{vaswani2017attention} and an MLP to $f$ and $\Delta_t$. In each convolution block, $\epsilon$ is linearly projected and added along the channel dimension of feature maps. \cite{gupta2023towards} additionally consider a scale-shift approach, where a linear projection of $\epsilon$ is multiplied element-wise along the channel dimension prior to addition along the channel dimension with a second linear projection of $\epsilon$~\citep{perez2018film}. \citet{gupta2023towards} refer to this approach as \textit{adaptive group normalization}, since the scale and shift are applied directly following normalization~\citep{nichol2021improved}. Since SineNet does not use pre-activation as in the \textsc{U-Net-Mod} considered by \cite{gupta2023towards}, wherein normalization and activation functions are applied prior to convolution layers, the scale and shift in SineNet are instead applied following the activation function. Nonetheless, we still refer to the scale-shift approach as adaptive group normalization for consistency with \cite{gupta2023towards}.
    % Since \cite{gupta2023towards} use pre-activation, that is apply normalization and activation prior to convolution, this scale and shift is applied following normalization, and thus, they refer to this approach as adaptive group normalization~\cite{nichol2021improved}. In contrast,       adapative group normalization, which involves both an element-wise multiplication and addition of linear projections of $\epsilon$ along the channel dimensions~\citep{nichol2021improved,perez2018film}, however, we do not consider this approach here.

    In Table~\ref{tab:cond}, we compare \textsc{SineNet-8} to \textsc{U-Net-Mod} using both the additive (\textsc{Add}) and adaptive group normalization (\textsc{AdaGN}) approaches on the conditional task.
    % Additionally, given the strong performance of \textsc{Dil-ResNet} on INS in Table~\ref{tab:res}, we compare to a conditional version referred to as \textsc{Dil-ResNet-Add}. 
    Models are trained on a train/valid/test split of 2,496/95/608, where all trajectories have 56 time steps and the buoyancies $f$ appearing in each split are distinct from those appearing in the remaining splits. Following~\cite{gupta2023towards}, we evaluate each of the models on 1-step prediction with $\Delta_t\in\{0.375, 0.75, 1.5, 3.0,6.0\}$. We additionally report errors on rollouts of length 10 with $\Delta_t=0.375$ beginning from all possible time steps in each trajectory. Consistent with \cite{gupta2023towards}, we find the \textsc{AdaGN} versions of SineNet and \textsc{U-Net-Mod} to outperform the \textsc{Add} versions. \textsc{SineNet8-AdaGN} is the top performer in all metrics. 

            \begin{table}[h]
            \caption{Results on \textsc{INS} for \textsc{SineNet-8}, \textsc{SineNet-Neural-ODE}, and \textsc{SineNet-8} with reduced time history $h$.}
            % , and \textsc{SineNet}-8 with a bottleneck between waves.}
            \begin{center}
            \label{tab:abl2}
                \begin{sc}
                    % \resizebox{\textwidth}{!}{
    \begin{tabular}{lccccccc}
        \toprule
        Method & \# Par. (M)  & 1-Step ($\%$) & Rollout ($\%$) \\
        \midrule
        SineNet-8 & 35.5 & 1.66 & 19.25\\
        SineNet-Neural-ODE & 12.7 & 1.83 & 19.88 \\
        SineNet-8, $h=1$ & 35.5 & \textbf{1.62} & \textbf{18.86} \\
        SineNet-8, $h=2$ & 35.5 & 1.64 & 19.07 \\
        SineNet-8, $h=3$ & 35.5 & \underline{1.62} & \underline{18.90} \\
        % SineNet-8-BottleNeck & 35.5 & 1.68 & 19.19 \\
        % \midrule
        % SineNet-8-128 & 28.3 & \textbf{1.41} & \textbf{17.15} & \textbf{0.82} & 2.15 & \textbf{0.93} & \textbf{1.66} \\
        % Deeper U-Net-8-128 & 20.1 & 1.46 & 17.22 & 0.84 & \textbf{1.92} & 1.33 & 2.51 \\
        \bottomrule
    \end{tabular}
                    % }
                \end{sc}
            \end{center}
        \end{table}

\subsection{Connection to Neural ODE}\label{sec:node}

    Neural ODE~\citep{chen2018neural} is a general neural framework for learning the mapping from an initial latent representation $\bm{h}(0)$ to an output latent representation $\bm{h}(1)$ by parameterizing the derivative of $\bm h$ with a neural network $v_\theta$ as 
    $$\frac{d\bm h(\tau)}{d\tau}=v_\theta(\bm h(\tau), \tau).$$
    To map from $\bm h(0)$ to $\bm h(1)$, the derivative can then be integrated as
    $$
    \bm h(1)=\bm h(0)+\int_0^1 v_\theta(\bm h(\tau), \tau)d\tau, 
    $$
    where the integral is approximated using a numerical integrator. \citet{chen2018neural} derive a method for backpropagating through the operations of an arbitrary numerical integrator in a stable and memory-efficient manner, enabling training of $v_\theta$. Their method includes the case of adaptive time-stepping integrators, which evaluate the derivative $\frac{d\bm h (\tau)}{d\tau}$ an adaptive number of times dependent on $\bm  h(\tau)$ to induce stability in the integration. This allows $v_\theta$ to be applied a variable number of times in mapping $\bm h(0)$ to $\bm h(1)$, thereby formulating a continuous-depth neural network.

    A number of common machine learning tasks fit in this framework, including image classification, image generation, and time series modeling. Taking $\bm h(\tau)$ as the latent solution $\bm x_{t+\tau}$ and $v_\theta$ as a U-Net frames the mapping from $\bm x_t$ to $\bm x_{t+1}$ as a neural ODE formulation of SineNet which we refer to as $\textsc{SineNet-Neural-ODE}$. We train and evaluate $\textsc{SineNet-Neural-ODE}$ on INS using the fifth order Dormand-Prince-Shampine numerical integrator, which is the default in the Neural-ODE PyTorch library~\citep{torchdiffeq}. Following the MNIST~\citep{lecun1998gradient} experiment from~\cite{chen2018neural}, we reduce the parameters of $\textsc{SineNet-Neural-ODE}$ roughly by a factor of 3 relative to SineNet. We incorporate the coordinate $\tau$ via a learned embedding vector added along the channel dimension in each convolution block of $v_\theta$ as done in the conditional task in Appendix~\ref{app:conditional}. 

\begin{wrapfigure}[15]{r}{0.45\linewidth}
% \begin{figure}[h]
    \vspace{-10pt}
    \centering
    \includegraphics[width=\linewidth]{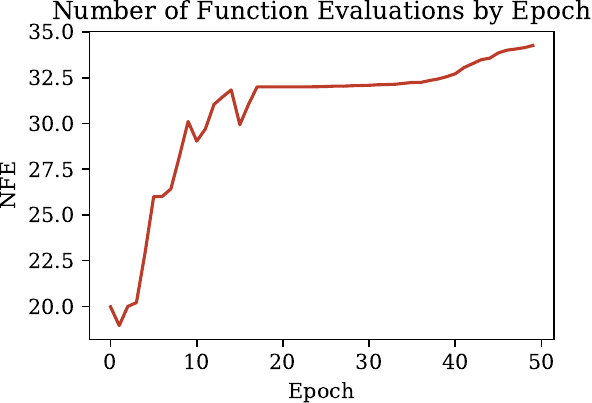}
    \caption{Average number of forward passes per batch by epoch when training \textsc{SineNet-Neural-ODE}. Consistent with~\citet{chen2018neural}, we find that the number of forward passes increases as training progresses.}
    \label{fig:nfe}
% \end{figure}
\end{wrapfigure}
    
    Although~\citet{chen2018neural} showed that $v_\theta$ could be optimized with space complexity independent of the number of evaluations, we instead train \textsc{SineNet-Neural-ODE} with direct back propagation to speed up training. As shown in Figure~\ref{fig:nfe}, the number of function evaluations for \textsc{SineNet-Neural-ODE} increases as training progresses, which is consistent with the findings of~\citet{chen2018neural}. Due to the high number of U-Net calls per forward pass, \textsc{SineNet-Neural-ODE} is expensive to train in terms of space and time despite the reduction in the number of parameters, as we show in Table~\ref{tab:ts}. We present test results for \textsc{SineNet-Neural-ODE} in Table~\ref{tab:abl2}, where we find its performance to be competitive with \textsc{SineNet-8}.
    
\subsection{Super-resolution analysis}\label{app:sr}

\begin{figure}[t]
    \centering
    \includegraphics[width=0.7\linewidth]{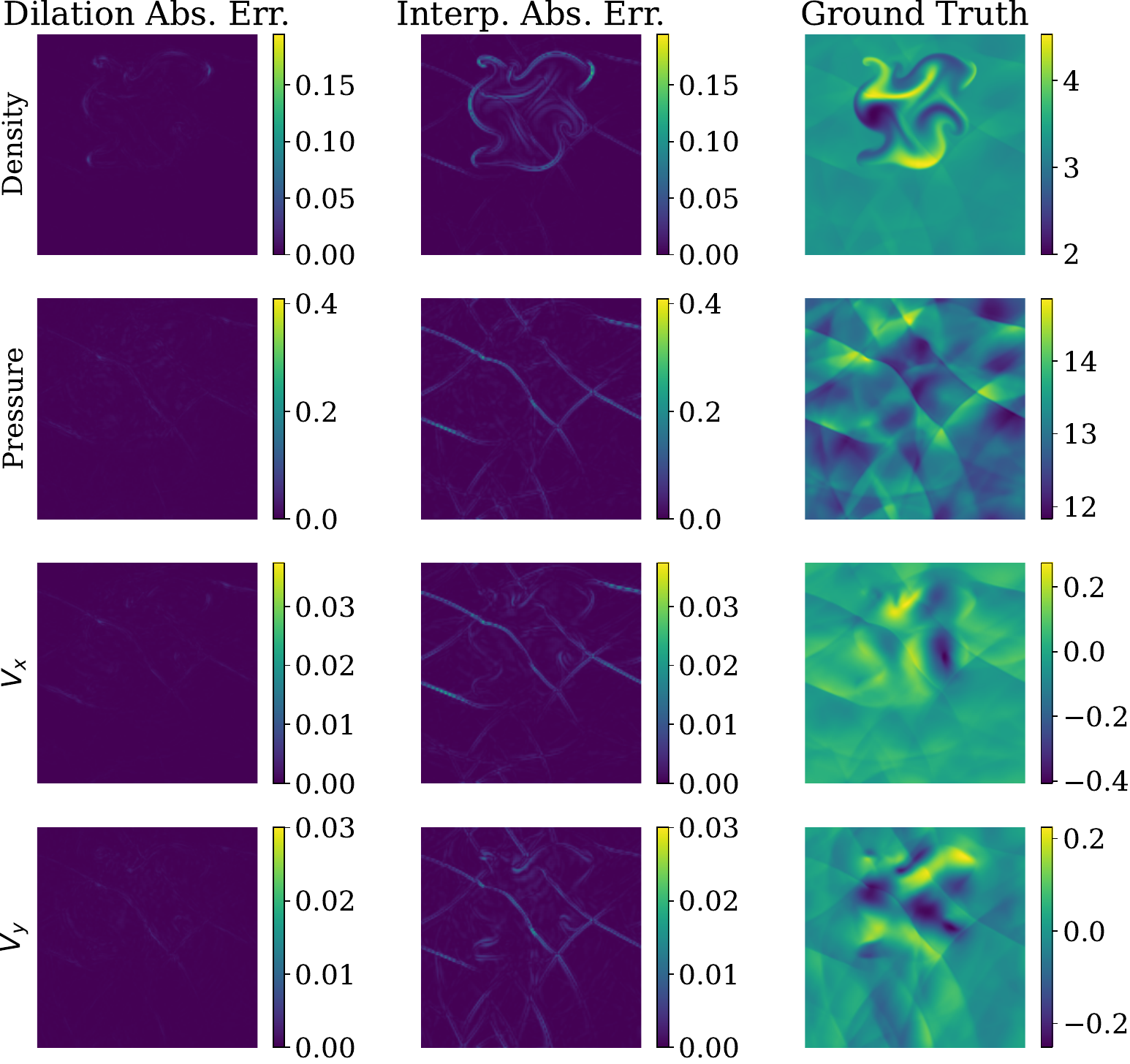}
    \caption{Absolute error for super-resolved one-step prediction on CNS using dilation and interpolation.}
    \label{fig:sr_vis}
\end{figure}

\begin{table}[h]
    \centering
    % \vspace{0.1in}
    \caption{Summary of rollout test error and one-step test error for the super-resolution task on 1,000 CNS trajectories downloaded from PDEBench~\citep{takamoto2022pdebench}. }\label{tab:sr}
    % \vspace{-5pt}
    % \small
    \begin{sc}
    \resizebox{\columnwidth}{!}{
     \begin{tabular}{lcccccc}
        \toprule
        & & \multicolumn{2}{c}{$128\times128$} & & \multicolumn{2}{c}{$512\times512$} \\
        Method & \# Par. (M) & 1-Step (\%) & Rollout (\%) & SR Method & 1-Step (\%) & Rollout (\%) \\
        \midrule
        \multirow{2}{*}{SineNet-8} & \multirow{2}{*}{35.5} & \multirow{2}{*}{\textbf{1.06}} & \multirow{2}{*}{\textbf{2.64}} & Dilation & \textbf{1.06} & \textbf{2.65} \\
         &  &  & & Interpolation & 2.44 & 3.54 \\
         \midrule
        F-FNO & 30.1 & 1.46 & 2.93 & Direct & 1.44 & 2.90 \\
        \bottomrule
    \end{tabular}
    }
    \end{sc}
    % \vspace{-5pt}
\end{table}

As discussed in Section~\ref{sec:neuralSolver}, neural operators~\citep{kovachki2021neural} aim to learn PDE solution operators independently of the resolution of the training data. This enables generalization beyond the discretization of the training data such that a trained neural operator can perform \textit{zero-shot super resolution}~\citep{li2021fourier}, wherein the task is to solve the PDE at a higher resolution than during training. As U-Nets, SineNets, and other CNN-based architectures learn their kernel functions on the same grid as the training data, they cannot perform this task directly, although recent work has adapted CNNs to the neural operator framework~\citep{raonic2023convolutional}. 

To perform super-resolution with SineNet, we consider two approaches: \textbf{interpolation} and \textbf{dilation}.

\textbf{Interpolation.} The input initial 10 time steps are downsampled to the training resolution of $128\times128$. SineNet then solves the PDE at the lower resolution, after which the solution is interpolated to the higher resolution. 

\textbf{Dilation.} Alternatively, SineNet can operate directly on the high resolution data using dilation. Convolutions, downsampling, and upsampling operations from the SineNet trained on $128\times128$ are all dilated by a factor of 4. 
Intuitively, the grid can be viewed to be divided using a checkerboard pattern, where each grid point interacts exclusively with other grid points that share the same type (color) as designated by this checkerboard arrangement.
This ensures that even at a higher resolution, each feature map grid point interacts only with grid points spaced equidistant to those it would interact with at the training resolution.
Although dilation is a standard operation in convolution layers, we highlight that it is crucial to also apply dilation in pooling (for downsampling) and in interpolation (for upsampling).
For example, in the $2\times 2$ pooling operation, instead of averaging and reducing over $2 \times 2$ regions of immediate neighboring grid points, we average and reduce over the 4 corner grid points of $5\times 5$ regions.
% the neighborhood of grid points that communicate with a given feature map's grid point is identical to the neighborhood during training  
% communication between grid points is constrained to the same resolution of 
% value the value of feature maps at each grid point is only influenced by values in the previous feature map that are spaced identically as at the training resolution.

We download 1,000 $512\times 512$ CNS trajectories from PDEBench~\citep{takamoto2022pdebench} and evaluate models trained on the $128\times 128$ CNS data presented in the main text. We compare to \textsc{F-FNO}, a neural operator which does not require interpolation or dilation to perform super-resolution. Results are presented in Table~\ref{tab:sr}. As in the main text, \textsc{SineNet-8} outperforms \textsc{F-FNO} at training resolution. At the higher resolution, interpolation introduces error, while dilation achieves nearly identical error to at training resolution. In Figure~\ref{fig:sr_vis}, we visualize the super-resolved errors using both approaches for one-step prediction on a randomly chosen example. As can be seen, unlike dilation, interpolation introduces errors in regions of high gradient. However, dilation is only applicable for super-resolving at integer multiples of the training resolution. 
For non-integer super-resolution multiples, similar techniques in deformable convolutions~\citep{dai2017deformable} could be considered to offset the operations to non-integer locations during testing.
Furthermore, all super-resolution approaches introduce error in settings where the higher resolution solutions contain high-frequency details not present at training resolution, or when numerical simulation at a higher resolution alters the behavior of the dynamics relative to training resolution, \textit{e.g.,} by introducing smaller eddies in a fluid simulation.

\subsection{Temporal evolution of rollout error}

\begin{figure}[t]
    \centering
    \includegraphics[width=\linewidth]{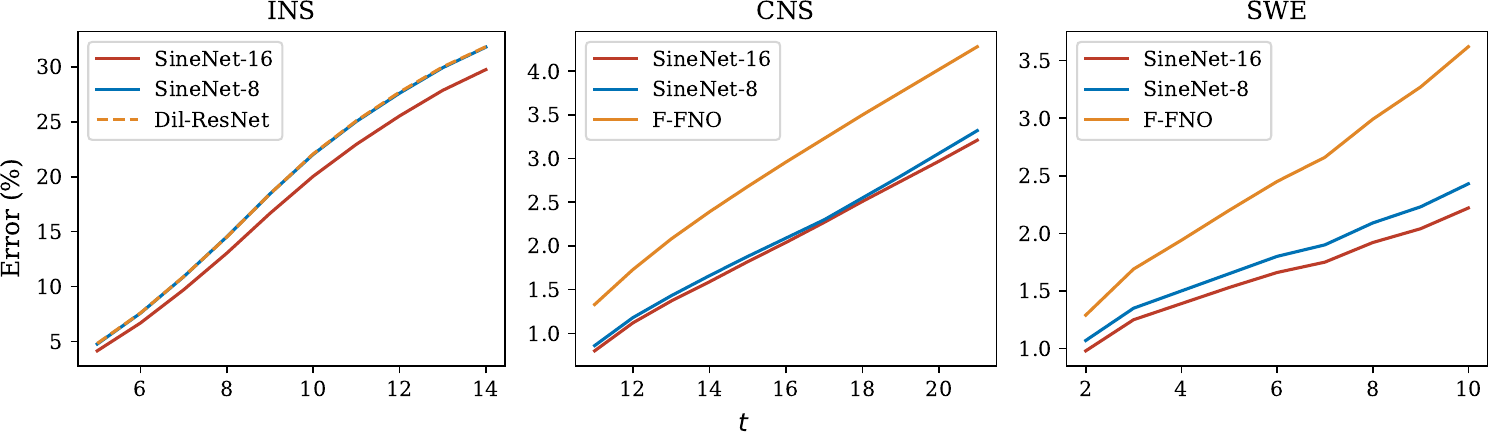}
    \caption{Evolution of rollout error for \textsc{SineNet-8}, \textsc{SineNet-16}, and the best baseline on each dataset.}
    \label{fig:rollerrs}
\end{figure}

In Figure~\ref{fig:rollerrs}, we visualize the evolution of the rollout error of SineNet against the best baseline on each of the considered datasets.

\subsection{Long-time prediction on CNS}\label{app:cns120}

To evaluate SineNet on longer time horizons, we generated 100 CNS trajectories with $T=120$ time steps. As opposed to the CNS task considered in the main text where solvers unroll trajectories for 11 steps given the initial 10 steps, we increase the time horizon by a factor of 10 in unrolling 110 steps given the initial 10 steps. Although the initially turbulent dynamics stabilize over the lengthy trajectory and therefore present a more stable prediction target, we evaluated models trained on the original CNS task such that beyond $t=21$, the dynamics were out-of-distribution.

\begin{wrapfigure}[18]{r}{0.45\linewidth}
% \begin{figure}[h]
    \vspace{-10pt}
    \centering
    \includegraphics[width=\linewidth]{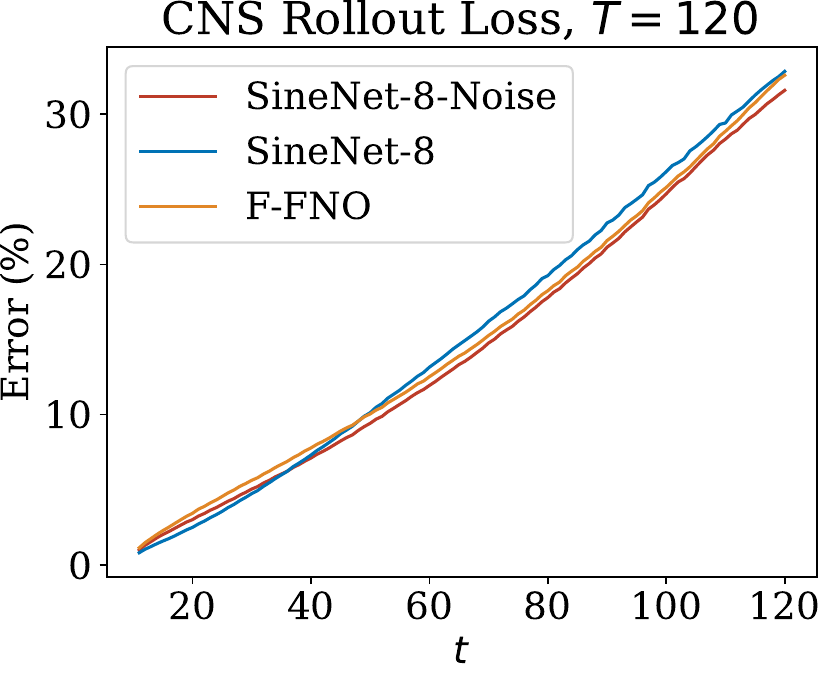}
    \caption{Rollout error averaged across 89 CNS trajectories of length $T=120$ for \textsc{SineNet-8-Noise}, \textsc{SineNet-8}, and \textsc{F-FNO}.}
    \label{fig:roll120}
% \end{figure}
\end{wrapfigure}

\begin{figure}[t]
    \centering
    \includegraphics[width=0.7\linewidth]{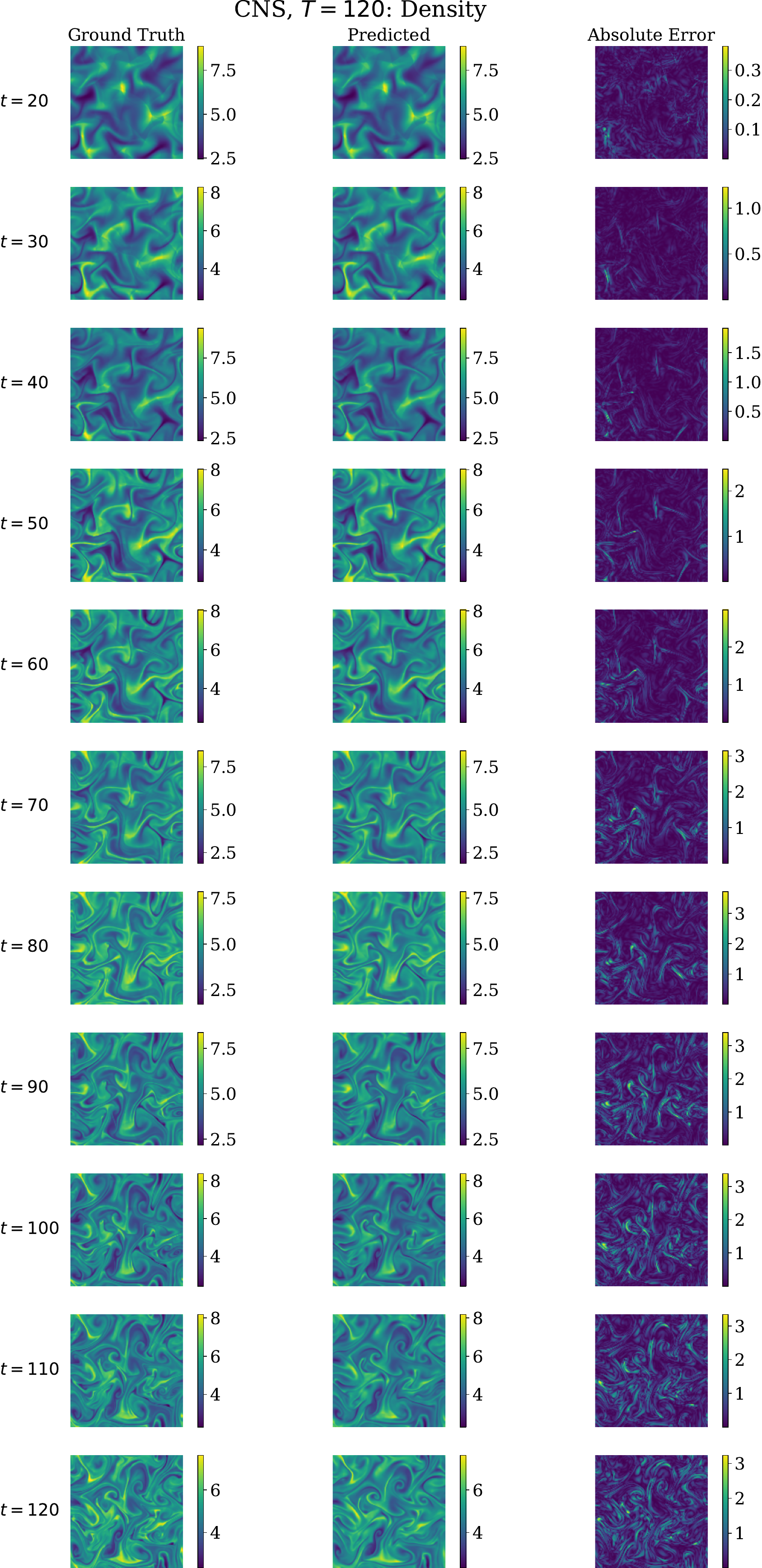}
        \caption{Compressible Navier-Stokes density field for $T=120$ downsampled to every $10$ time steps.}
    \label{fig:CNSdensity_pred120}
\end{figure}

\begin{figure}[t]
    \centering
    \includegraphics[width=0.7\linewidth]{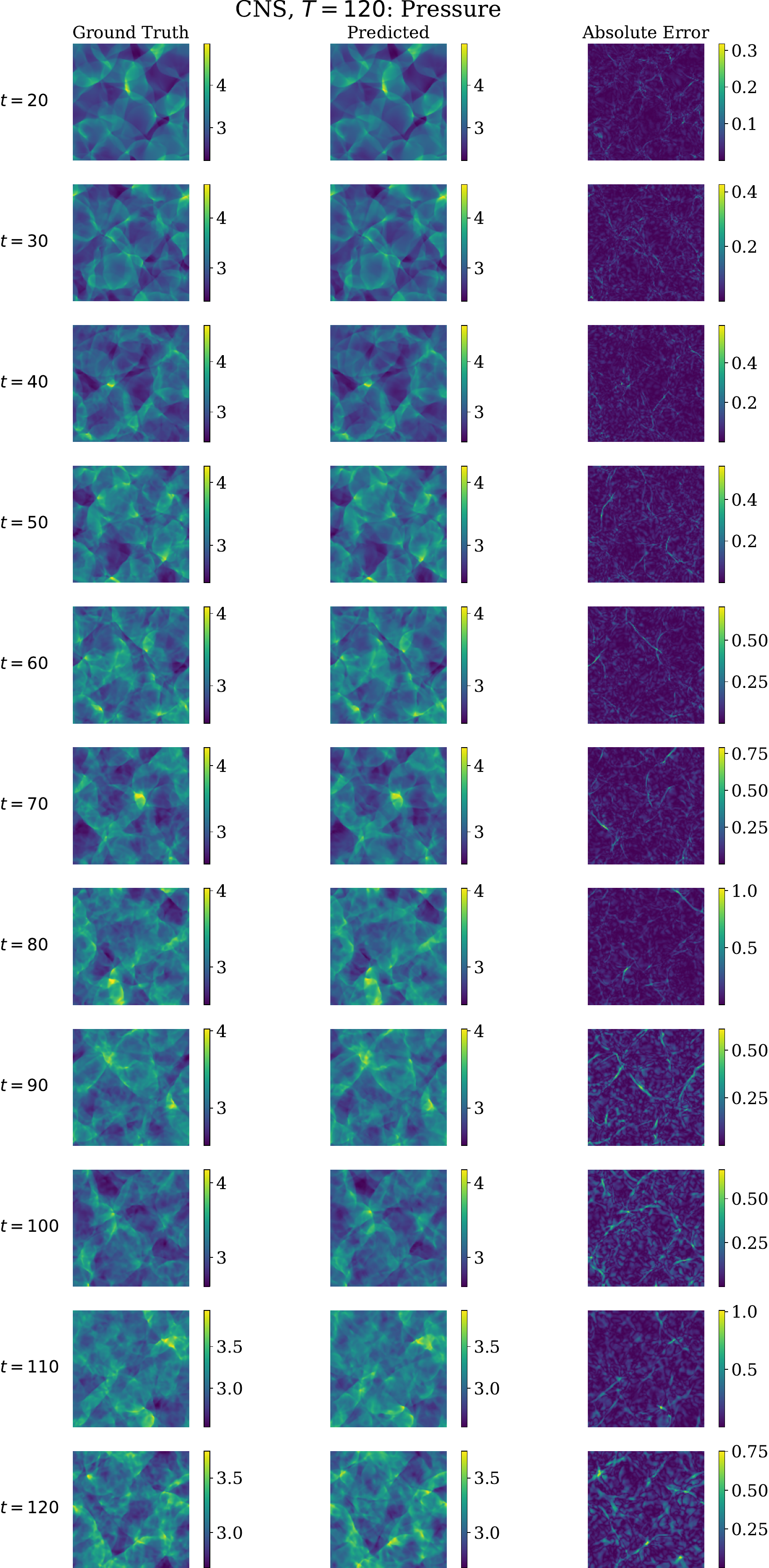}
        \caption{Compressible Navier-Stokes pressure field for $T=120$ downsampled to every $10$ time steps.}
    \label{fig:CNSpressure_pred120}
\end{figure}

\begin{figure}[t]
    \centering
    \includegraphics[width=0.7\linewidth]{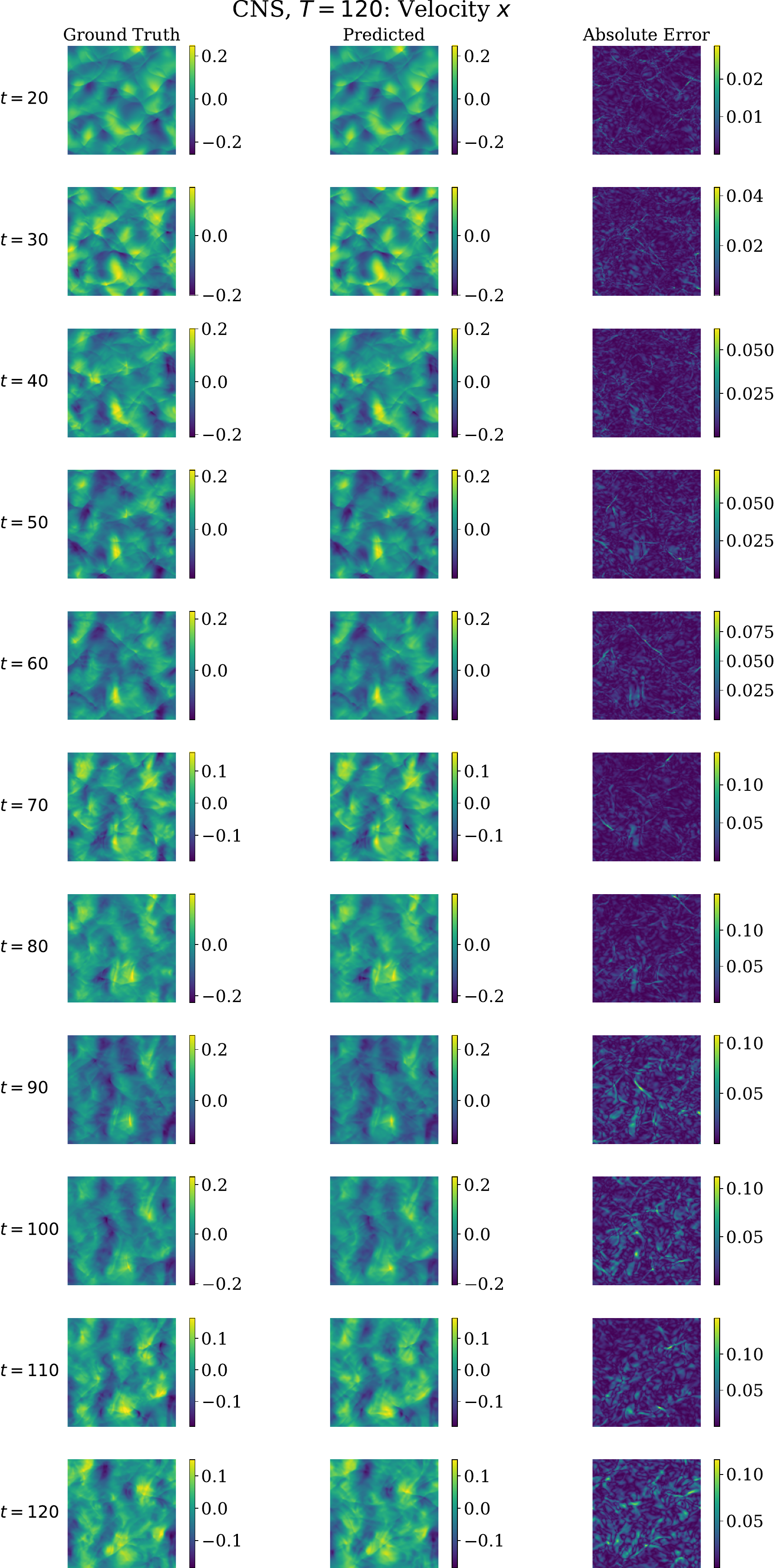}
    \caption{Compressible Navier-Stokes velocity $x$ component for $T=120$ downsampled to every $10$ time steps.}
    \label{fig:CNSvelx_pred120}
\end{figure}

\begin{figure}[t]
    \centering
    \includegraphics[width=0.7\linewidth]{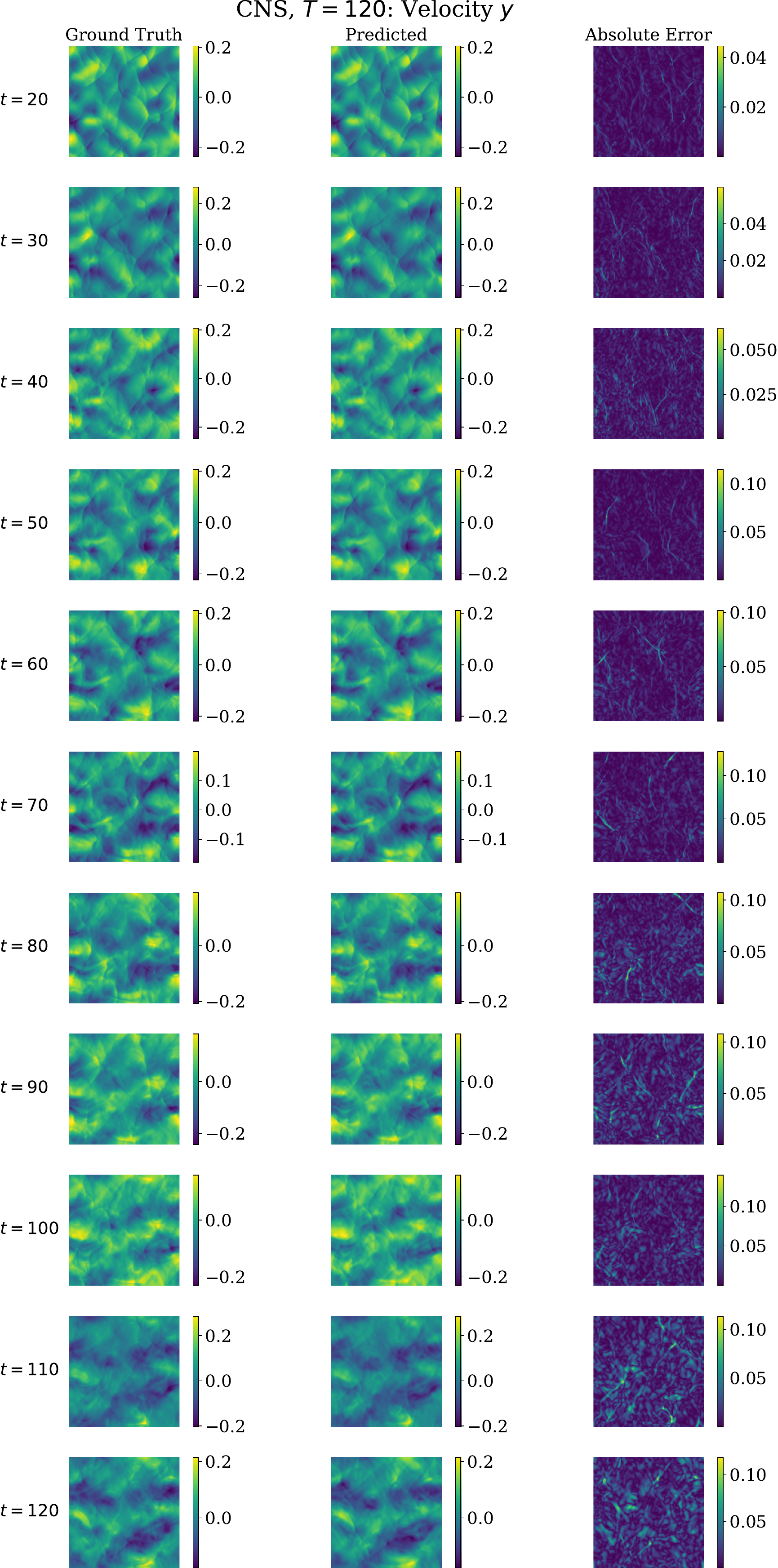}
    \caption{Compressible Navier-Stokes velocity $y$ component for $T=120$ downsampled to every $10$ time steps.}
    \label{fig:CNSvely_pred120}
\end{figure}

One approach to increase long-term stability is noise injection~\citep{sanchez2020learning,stachenfeld2022learned}, where 0-mean Gaussian noise is added to training inputs to simulate noise encountered by the solver during rollouts due to errors in previous predictions. Recent work by \cite{lippe2023pde} extended this approach to consider multiple levels of noise that the model is trained to remove, which shares a connection with denoising diffusion probabilistic models~\citep{ho2020denoising}. \citet{tran2023factorized} found noise injection to be important for stable training of F-FNO, which we employed in training F-FNO for our tasks. For this longer rollout task, we additionally evaluated a \textsc{SineNet-8-Noise} trained on the 21-step CNS data with the noise level $\sigma=0.01$, which is the same as \textsc{F-FNO}.

We present results in Table~\ref{tab:CNS120}. We find a subset of the generated trajectories led to poor performance across all models, even causing several of the baselines to diverge entirely and produce solutions with greater than 100\% error in 11/100 cases. The cause is unclear, although the initial density and/or pressure fields for each of these trajectories have small mean values relative to the remaining trajectories, potentially creating out-of-distribution dynamics. We therefore present results both with and without these 11 trajectories. \textsc{SineNet-8-Noise} has the lowest rollout error in both cases, and is more robust to the 11 difficult cases than the remaining models. 

In Figure~\ref{fig:roll120}, we present the rollout error by time step for~\textsc{SineNet-8} trained with and without noise, as well as \textsc{F-FNO}. As can be seen, noise injection has a regularizing effect that causes the predictions of \textsc{SineNet-8} to have better errors for short time horizons than models trained with noise injection. However, at longer time horizons, models trained with noise injection surpass \textsc{SineNet-8} due to their robustness to the accumulation of rollout error. In Figures~\ref{fig:CNSdensity_pred120}-\ref{fig:CNSvely_pred120}, we visualize a randomly selected trajectory predicted by \textsc{SineNet-8-Noise} on this data.

        \begin{table}[h]
            \caption{Summary of rollout test errors on CNS with $T=120$. \textsc{Rollout-89} corresponds to the dataset with the 11 trajectories resulting in a baseline error greater than 100\% removed, while \textsc{Rollout-100} is for the full dataset. `--' indicates a rollout error greater than 100\%.}
            \vspace{-5pt}
            \begin{center}
            \label{tab:CNS120}
                \begin{sc}
                    % \resizebox{\textwidth}{!}{
    \begin{tabular}{lccc}
        \toprule
        Method & \# Par. (M) & Rollout-100 ($\%$) & Rollout-89 ($\%$) \\
        \midrule
        SineNet-8-Noise & 35.5 & \textbf{15.94} & \textbf{14.62} \\
        SineNet-8 & 35.5 & \underline{27.97} & 15.44 \\
        F-FNO & 30.1 & -- & \underline{15.15} \\
        Dil-ResNet & 16.7 & -- & 23.44 \\
        U-Net-128 & 135.1 & 29.97 & 15.25 \\
        U-Net-Mod & 144.3 & -- & 18.95 \\
        \bottomrule
    \end{tabular}

                    % }
                \end{sc}
            \end{center}
        \end{table}    

\subsection{Ablation on number of conditioning steps}

    In Table~\ref{tab:abl2}, we ablate the number of historical conditioning steps used on INS. We find that $h=1$ gives the lowest test error, followed by $h=3$.
% Surprisingly, we found that the test error decreases when less historical steps are used, and conditioning on only 1 step gives the best result. This suggests that the majority of the information can be effectively captured in the most recent time step. 
That $h=1$ performs as well or better than $h>1$ is in line with findings reported by~\cite{tran2023factorized}. In our experiments, we instead adhere to the number of historical steps used by the respective benchmarks on each dataset  
% In the main paper, we mainly follow the setting in the benchmark papers so that the results can be compared to previous work. As a result, we still 
 % therefore use more historical steps 
 (4 for INS, 10 for CNS and 2 for SWE). However, we believe that 
 % how to 
 principled approaches for choosing the number of conditioning steps, as well as how to 
 % efficiently use these 
 incorporate them into model predictions, is an interesting topic for future research.

\subsection{Wave bottleneck}

            \begin{table}[h]
            \caption{Results on \textsc{INS}, \textsc{CNS}, and \textsc{SWE} for \textsc{SineNet-8} and \textsc{SineNet-8-BottleNeck}.}
            \vspace{-7pt}
            \begin{center}
            \label{tab:bottle}
                \begin{sc}
                    \resizebox{\textwidth}{!}{
    \begin{tabular}{lcccccccc}
        \toprule
        & & \multicolumn{2}{c}{INS} & \multicolumn{2}{c}{CNS} & \multicolumn{2}{c}{SWE} \\%\cline{3-6}
        Method & \# Par. (M)  & 1-Step ($\%$) & Rollout ($\%$) & 1-Step ($\%$) & Rollout ($\%$) & 1-Step ($\%$) & Rollout ($\%$)\\
        \midrule
        SineNet-8 & 35.5 & \textbf{1.66} & {19.25} & \textbf{0.93} & \textbf{2.10} & \textbf{1.02} & \textbf{1.78} \\
        SineNet-8-BottleNeck & 35.5 & 1.68 & \textbf{19.19} & 1.20 & 3.14 & 1.63 & 2.51 \\
        % \midrule
        % SineNet-8-128 & 28.3 & \textbf{1.41} & \textbf{17.15} & \textbf{0.82} & 2.15 & \textbf{0.93} & \textbf{1.66} \\
        % Deeper U-Net-8-128 & 20.1 & 1.46 & 17.22 & 0.84 & \textbf{1.92} & 1.33 & 2.51 \\
        \bottomrule
    \end{tabular}

                    }
                \end{sc}
                            \vspace{-10pt}
            \end{center}
        \end{table}

    In addition to the dual processing mechanism which we discuss in Section~\ref{sec:dual} and ablate with \textsc{SineNet-8-Entangled} in Section~\ref{sec:ablation}, a primary difference between the waves $V_k$ comprising SineNet and a conventional U-Net is the encoder and decoder present in conventional U-Nets ($P$ and $Q$ in Equations~\ref{eq:down} and \ref{eq:up}). SineNet instead maintains a high-dimensional representation between waves. We ablate this design choice with \textsc{SineNet-8-BottleNeck}, which decodes the latent solution $\bm{x}_{t+\Delta_{k}}$ output by $V_k$ to the lower-dimensional base space before encoding back to the latent space for input to $V_{k+1}$. Aside from the dual processing mechanism, this renders the architecture of each wave $V_k$ closer to that of a conventional U-Net, however, it creates a bottleneck between waves.

    % In Table~\ref{tab:abl2}, we present results for \textsc{SineNet-8-BottleNeck} evaluated on INS. Relative to \textsc{SineNet-8}, the bottleneck causes a slight reduction in the 1-step error. This in turn has a regularizing effect on the rollout loss, which improves over the rollout loss of \textsc{SineNet-8}. We visualize the decoded feature maps between each wave in Figure~\ref{fig:bottle_vis}. 
    
    In Table~\ref{tab:bottle}, we present results for \textsc{SineNet-8-BottleNeck} evaluated on INS, CNS, and SWE. On all datasets, \textsc{SineNet-8} outperforms \textsc{SineNet-8-BottleNeck} in terms of 1-step error. While \textsc{SineNet-8} has a lower rollout error than \textsc{SineNet-8-BottleNeck} on CNS and SWE by a substantial margin, \textsc{SineNet-8-BottleNeck} tops the rollout error of \textsc{SineNet-8} on INS.
    % Relative to \textsc{SineNet-8}, the bottleneck causes a slight reduction in the 1-step error on both datasets. For INS, this in turn has a regularizing effect on the rollout loss, which improves over the rollout loss of \textsc{SineNet-8}. By contrast, on SWE, the rollout loss for \textsc{SineNet-8-BottleNeck} is more than 40\% greater than that for \textsc{SineNet-8}. 
    
    % The mixed results on these datasets could be due to the difficult initial time steps in the INS rollouts relative to the remaining time steps. As can be seen in Figures~\ref{fig:INSdensity_pred}-\ref{fig:INSvely_pred}, velocities in earlier timesteps are greater than those in later timesteps, resulting in dynamics which are faster-evolving and with a greater degree of local variability. This could be a primary factor contributing to the gap between the 1-step error, which is averaged over all possible one-step pairs ranging from the beginning to the end of the trajectory, and the rollout error on the INS data. This can be quantified by comparing the 1-step loss on INS to the average error at the beginning of the trajectory visualized in Figure~\ref{fig:roll120}, where it can be seen that the one-step loss starting from the beginning of the trajectory is more than twice the one-step loss averaged over all possible starting position

    The mixed results on these datasets could be due to the difficulty of the initial time steps in the INS rollouts relative to the remaining time steps. Compared to the 1-step error of 1.66 for \textsc{SineNet-8} averaged over all possible time steps in INS trajectories, Figure~\ref{fig:rollerrs} shows that the errors starting from the beginning of the trajectory are over twice as high. This is likely due to the large velocities in earlier timesteps creating dynamics which are faster-evolving and with a greater degree of local variability than at later timesteps, which is exemplified in Figures~\ref{fig:INSdensity_pred}-\ref{fig:INSvely_pred}. As a result of larger errors early on in predicted trajectories, as well as the accumulation of error through autoregressive rollout, there is a substantially larger gap between 1-step errors and rollout errors on INS. We hypothesize that the bottleneck serves as a form of regularization which decreases performance in terms of 1-step prediction, but increases robustness to the difficult initial steps in INS. Performing a similar analysis for CNS and SWE, the difficulty of the initial timesteps instead appear similar to the remaining timesteps on both datasets. Thus, the regularization is not beneficial as before, and in fact leads to a performance drop, likely due to the information bottleneck between waves.

\begin{figure}[t]
    \centering
    \includegraphics[height=0.9\textheight]{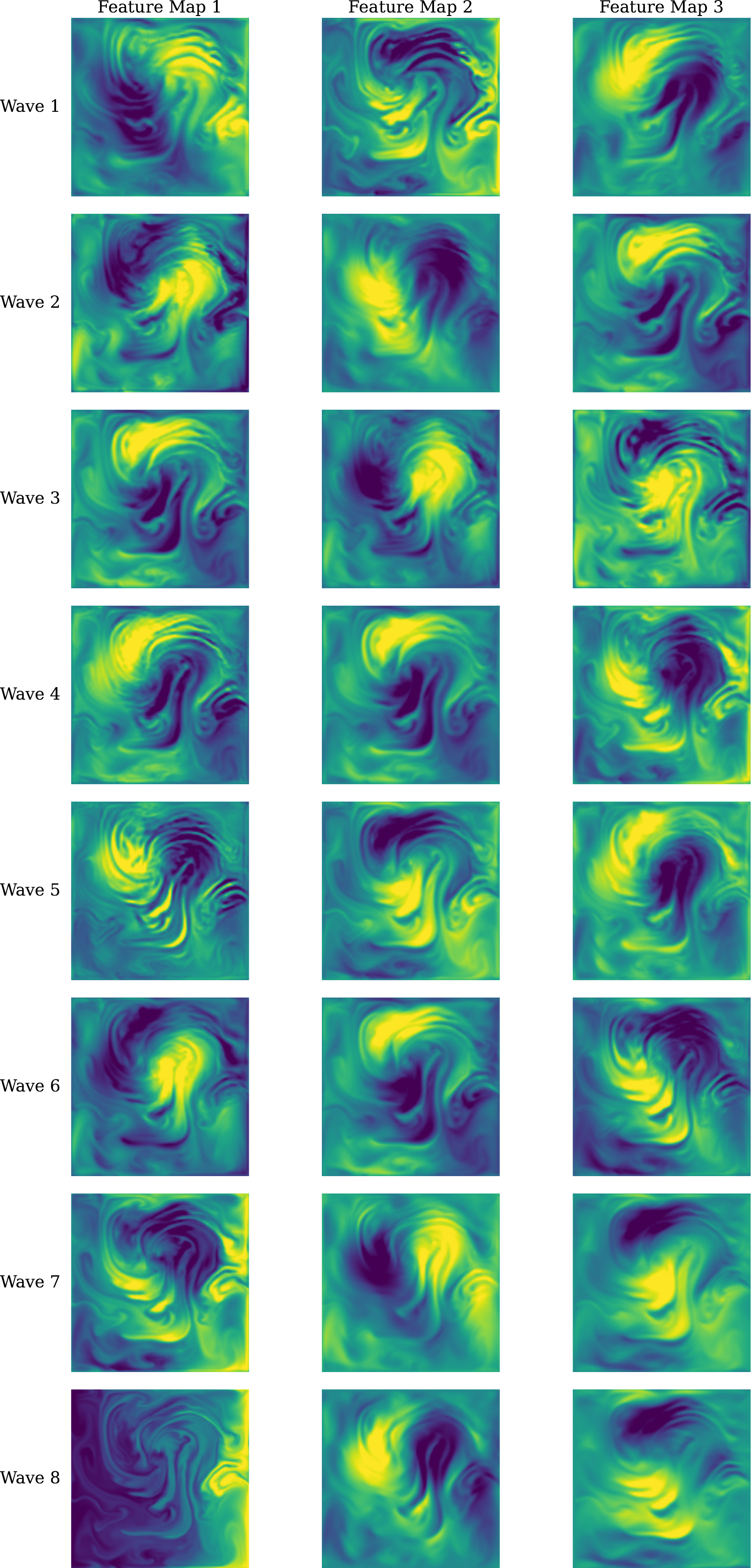}
    \caption{Visualization of \textsc{SineNet-8-BottleNeck} decoded feature maps between each wave on INS. The Wave 8 feature maps are the predicted particle concentration and velocity field.}
    \label{fig:bottle_vis}
\end{figure}

\begin{figure}[t]
    \centering
    \includegraphics[height=0.9\textheight]{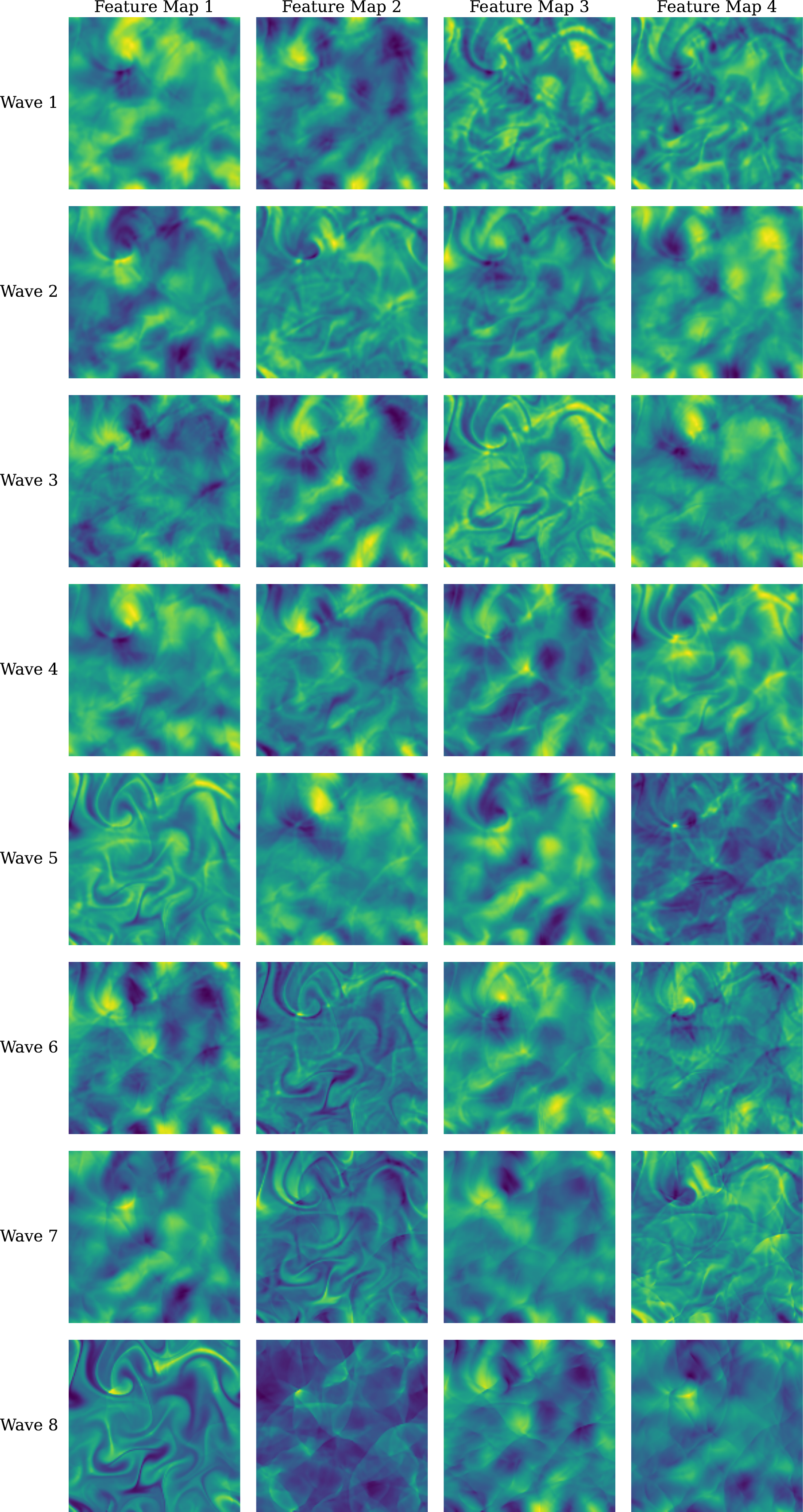}
    \caption{Visualization of \textsc{SineNet-8-BottleNeck} decoded feature maps between each wave on CNS. The Wave 8 feature maps are the predicted density, pressure and velocity fields.}
    \label{fig:bottle_vis_CNS}
\end{figure}

\begin{figure}[t]
    \centering
    \includegraphics[height=0.9\textheight]{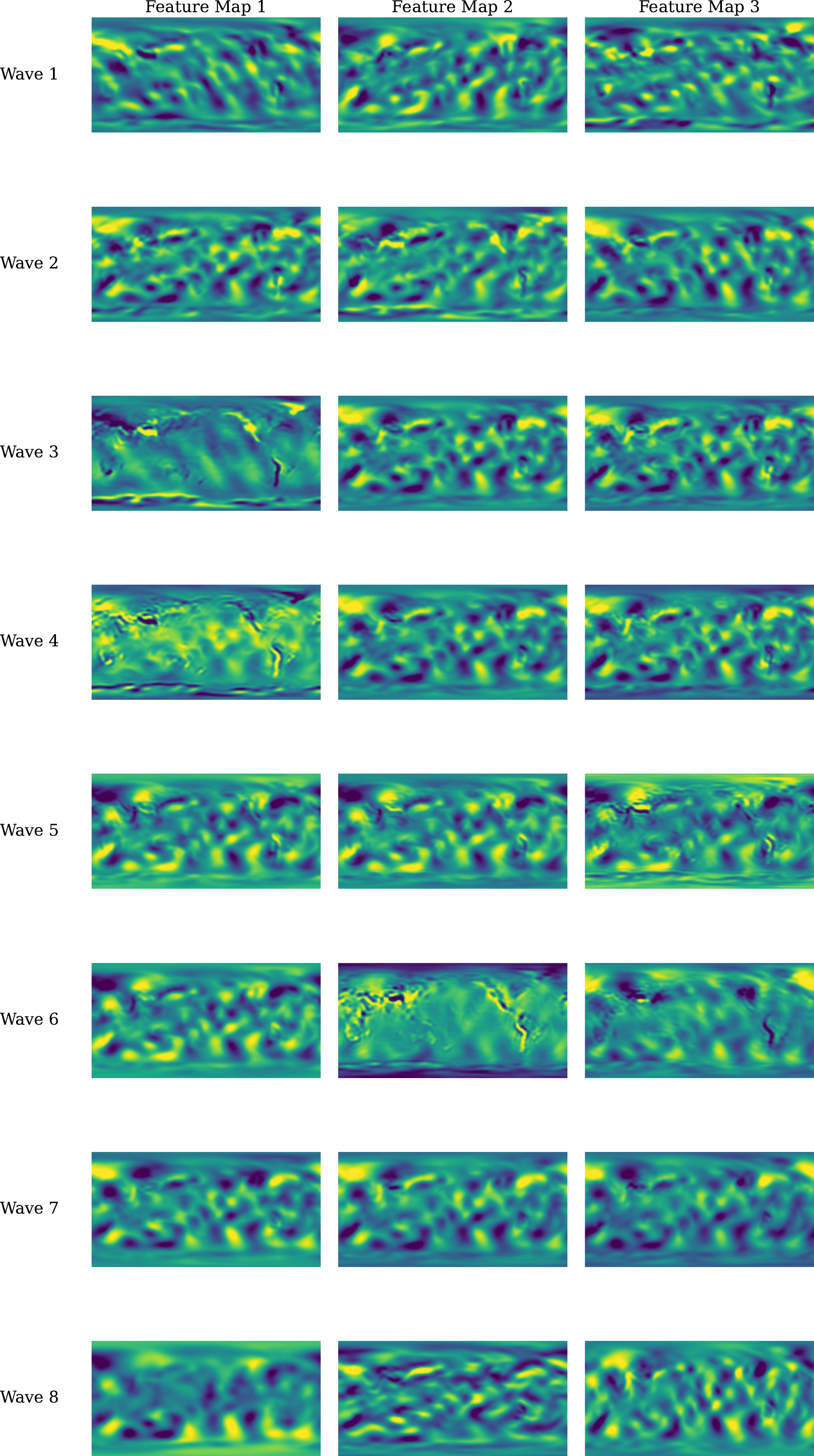}
    \caption{Visualization of \textsc{SineNet-8-BottleNeck} decoded feature maps between each wave on SWE. The Wave 8 feature maps are the predicted pressure and velocity field.}
    \label{fig:bottle_vis_SWE}
\end{figure}

    We visualize the decoded feature maps between each wave for \textsc{SineNet-8-BottleNeck} on INS, CNS, and SWE in Figures~\ref{fig:bottle_vis}, \ref{fig:bottle_vis_CNS}, and~\ref{fig:bottle_vis_SWE}, respectively. Interestingly, as SWE is a global weather forecasting task, we can see the outline of the world map in the visualized feature maps (\emph{e.g.,} Feature Map 1 of Wave 3 in Figure~\ref{fig:bottle_vis_SWE}), which implies that the primary objective of these feature maps is for modeling the evolution of dynamics about the boundaries of continents, potentially encouraged by the information compression in the wave bottleneck.
    % we observe in Figure~\ref{fig:bottle_vis_SWE} that 
    % which potentially implies 
    % that forcing the model to compress information between waves results in feature maps whose primary objective is for modeling the evolution of dynamics about the boundaries of continents. 
    % The presence of these land masses add a layer of complexity not present in the NS task which may have additionally contributed to the observed detriment of the bottleneck for SWE. 

\subsection{\textsc{SineNet}-$K$ Results}\label{app:numres}

In Table~\ref{tab:numres}, we present numerical results for the SineNets with varying $K$ visualized in Figure~\ref{fig:k_abl}.

        \begin{table}[h]
            \caption{Summary of rollout test error and one-step test error for SineNet with varying $K$.}
            \vspace{-7pt}
            % \caption{Test errors. The best performance is shown in bold and the second best is underlined.}
            \begin{center}
            \label{tab:numres}
                \begin{sc}
                    \resizebox{\textwidth}{!}{
    \begin{tabular}{lccccccc}
        \toprule
        & & \multicolumn{2}{c}{INS} & \multicolumn{2}{c}{CNS} & \multicolumn{2}{c}{SWE} \\%\cline{3-6}
        Method & \# Par. (M) & 1-Step ($\%$) & Rollout ($\%$) & 1-Step ($\%$) & Rollout ($\%$) & 1-Step ($\%$) & \textsc{Rollout ($\%$)} \\
        \midrule
        SineNet-2 & 35.5 & 2.38 & 23.71 & 1.56 & 4.98 & 1.38 & 2.54 \\
        SineNet-4 & 35.5 & 1.91 & 21.06 & 1.26 & 2.63 & 1.13 & 1.97 \\
       SineNet-6 & 35.5 & 1.73 & 19.58 & 1.12 & 2.40 & 1.10 & 1.91 \\
        SineNet-8 & 35.5 & 1.66 & 19.25 & 0.93 & 2.10 & 1.02 & 1.78 \\
        SineNet-10 & 35.5 & 1.58 & 18.68 & 0.94 & 2.24 & 0.97 & 1.73 \\
        SineNet-12 & 35.5 & 1.53 & 18.09 & 0.93 & 2.12 & 0.92 & 1.64 \\
        SineNet-14 & 35.4 & \underline{1.49} & \underline{17.66} & \underline{0.89} & \textbf{1.97} & \underline{0.94} & \underline{1.66} \\
        SineNet-16 & 35.5 & \textbf{1.46} & \textbf{17.63} & \textbf{0.87} & \underline{2.04} & \textbf{0.92} & \textbf{1.64}
        \\
        \bottomrule
    \end{tabular}

                    }
                \end{sc}
            \vspace{-11pt}
            \end{center}
        \end{table}

\color{black}
\captionsetup{
    font={color=black},
    textfont={color=black}
}

\section{Dataset and prediction visualization}\label{app:pred_vis}

\begin{figure}[t]
    \centering
    \includegraphics[width=0.7\linewidth]{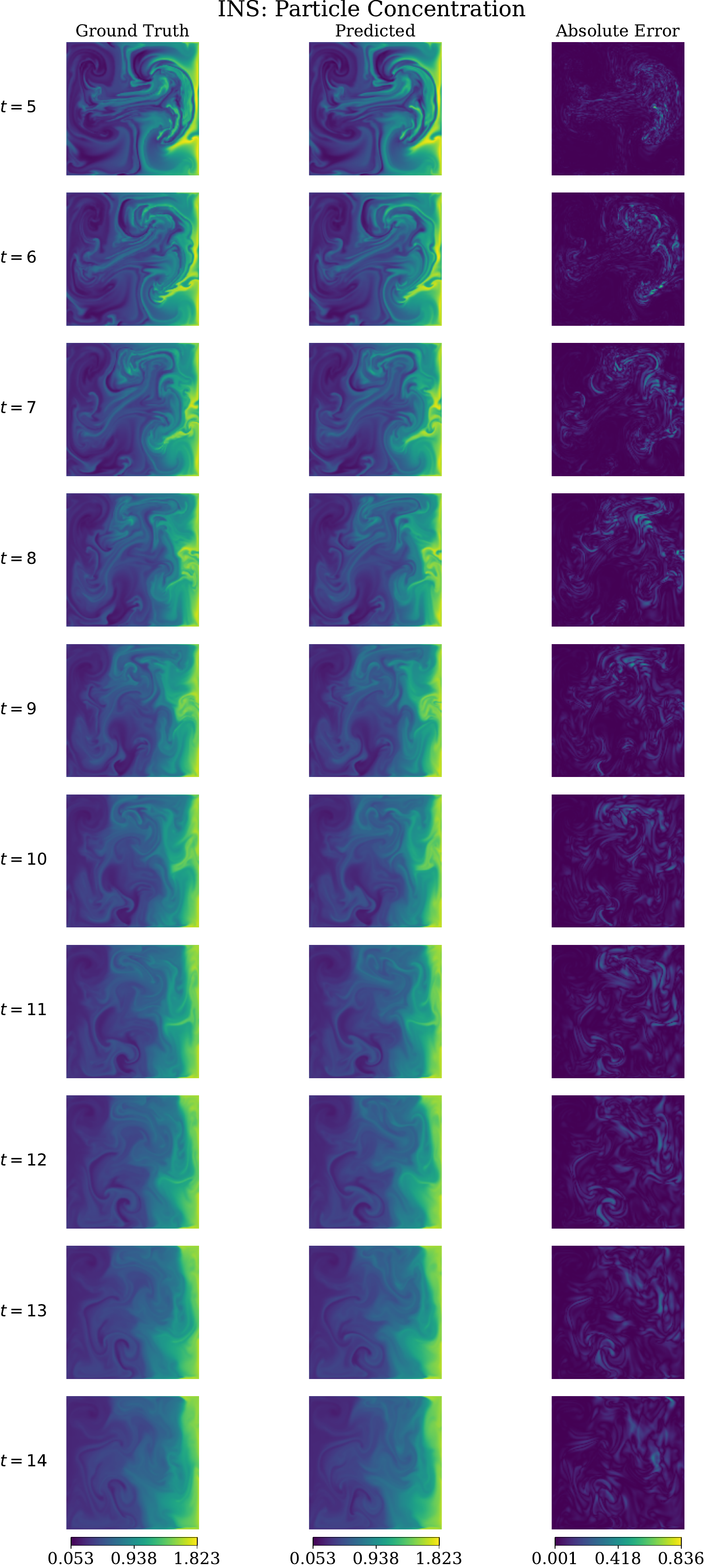}
        \caption{Incompressible Navier-Stokes particle concentration.}
    \label{fig:INSdensity_pred}
\end{figure}

\begin{figure}[t]
    \centering
    \includegraphics[width=0.7\linewidth]{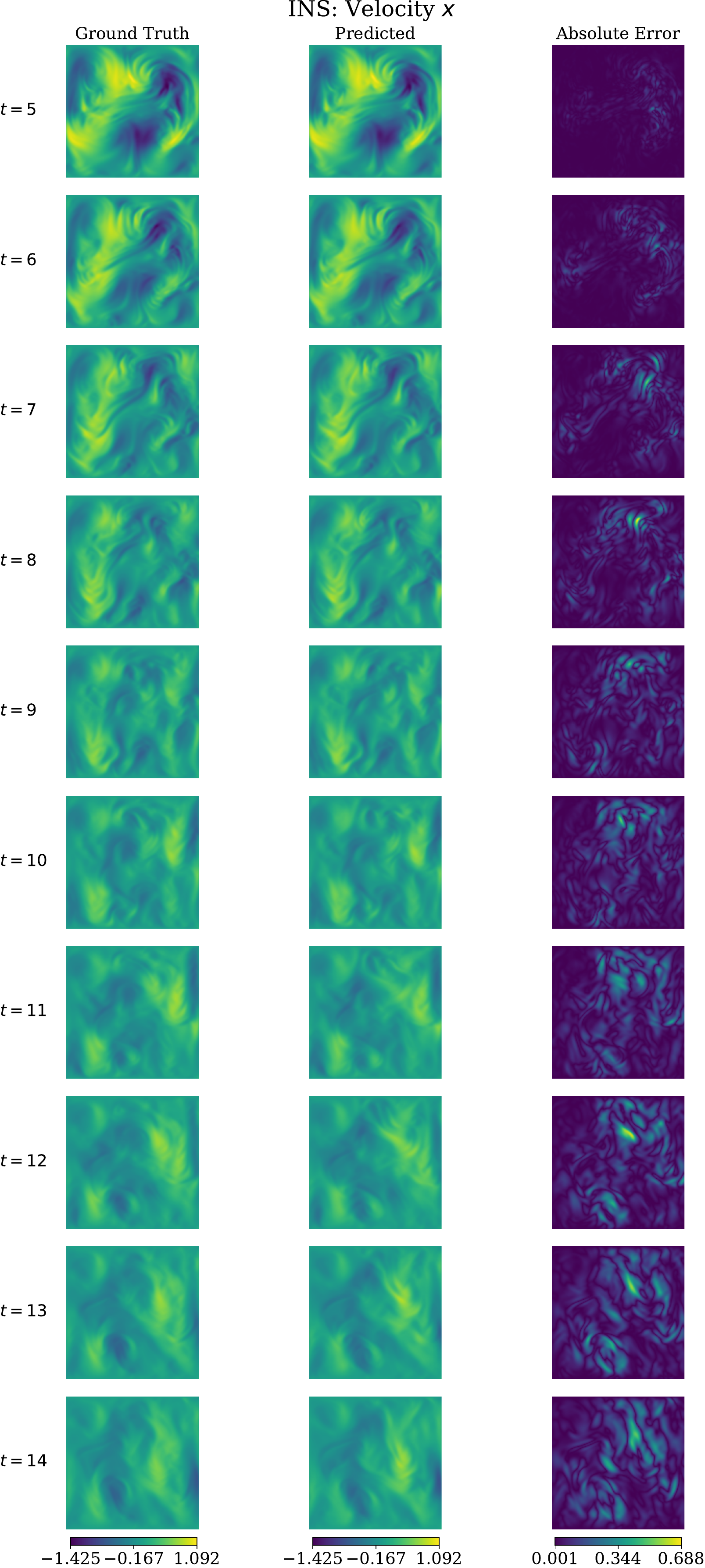}
    \caption{Incompressible Navier-Stokes velocity $x$ component.}
    \label{fig:INSvelx_pred}
\end{figure}

\begin{figure}[t]
    \centering
    \includegraphics[width=0.7\linewidth]{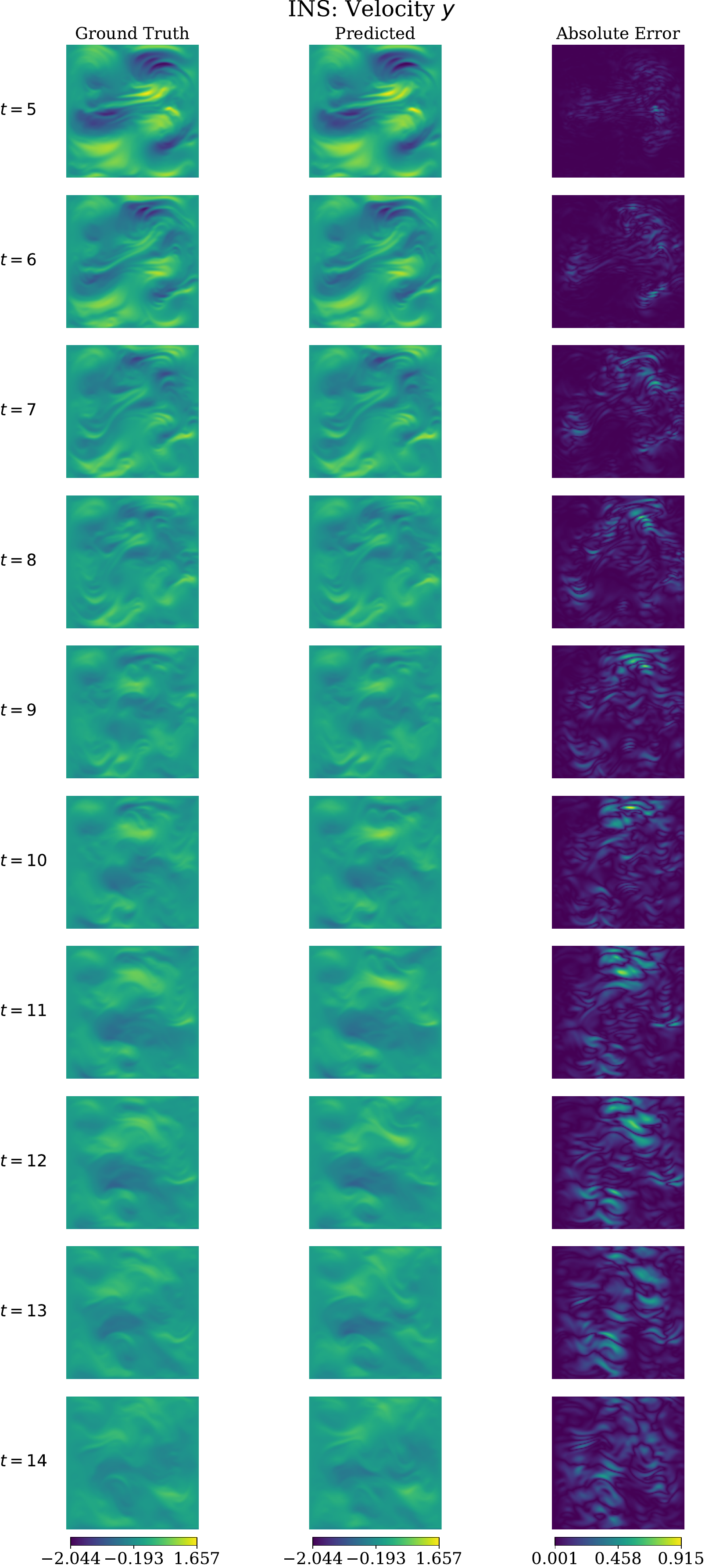}
    \caption{Incompressible Navier-Stokes velocity $y$ component.}
    \label{fig:INSvely_pred}
\end{figure}

\begin{figure}[t]
    \centering
    \includegraphics[width=0.7\linewidth]{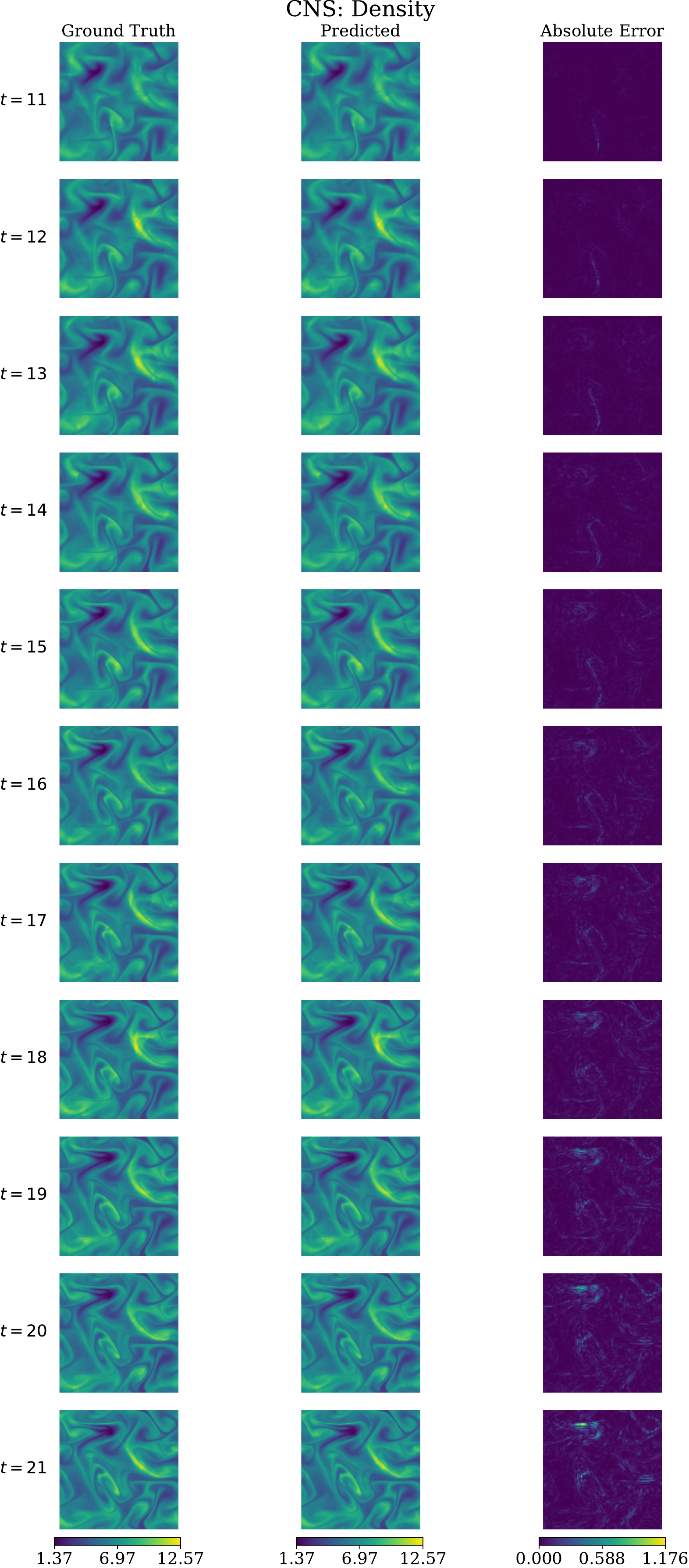}
        \caption{Compressible Navier-Stokes density field.}
    \label{fig:CNSdensity_pred}
\end{figure}

\begin{figure}[t]
    \centering
    \includegraphics[width=0.7\linewidth]{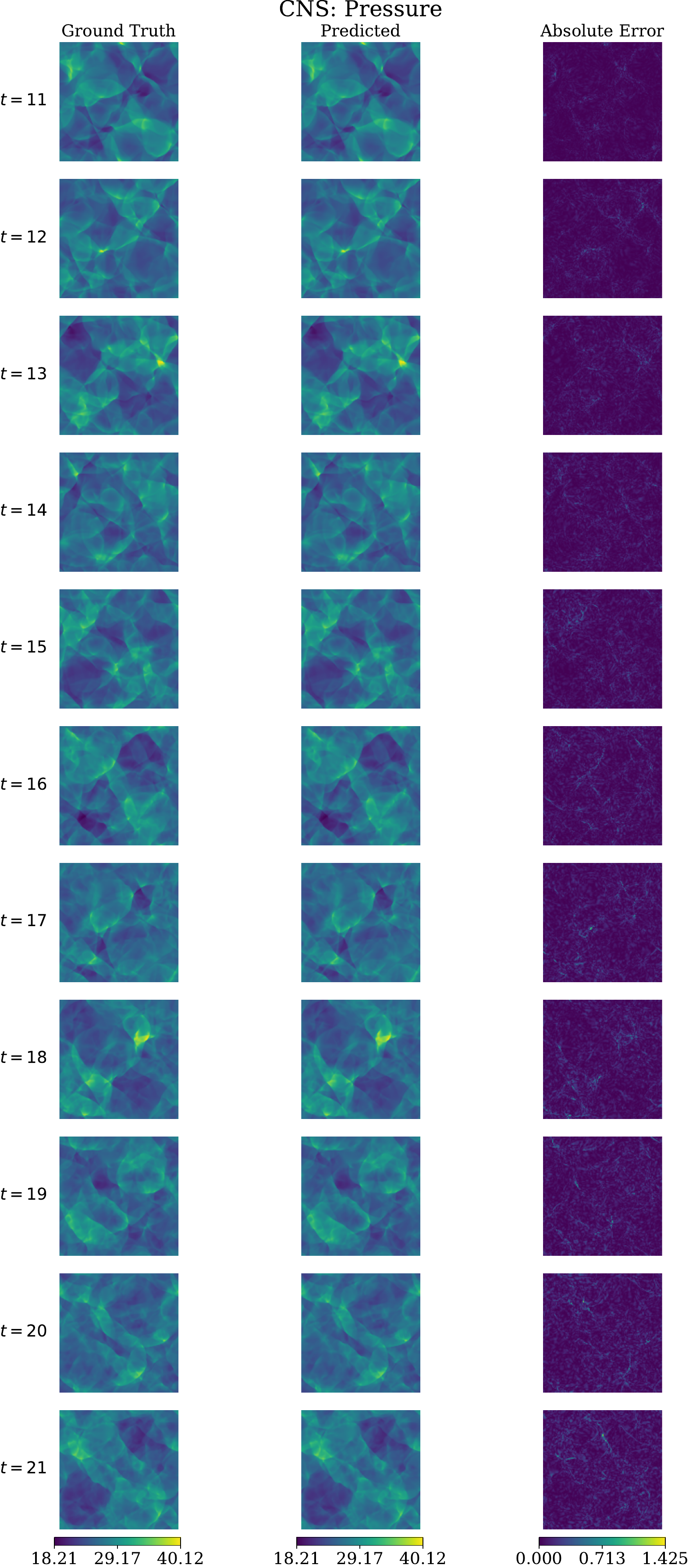}
        \caption{Compressible Navier-Stokes pressure field.}
    \label{fig:CNSpressure_pred}
\end{figure}

\begin{figure}[t]
    \centering
    \includegraphics[width=0.7\linewidth]{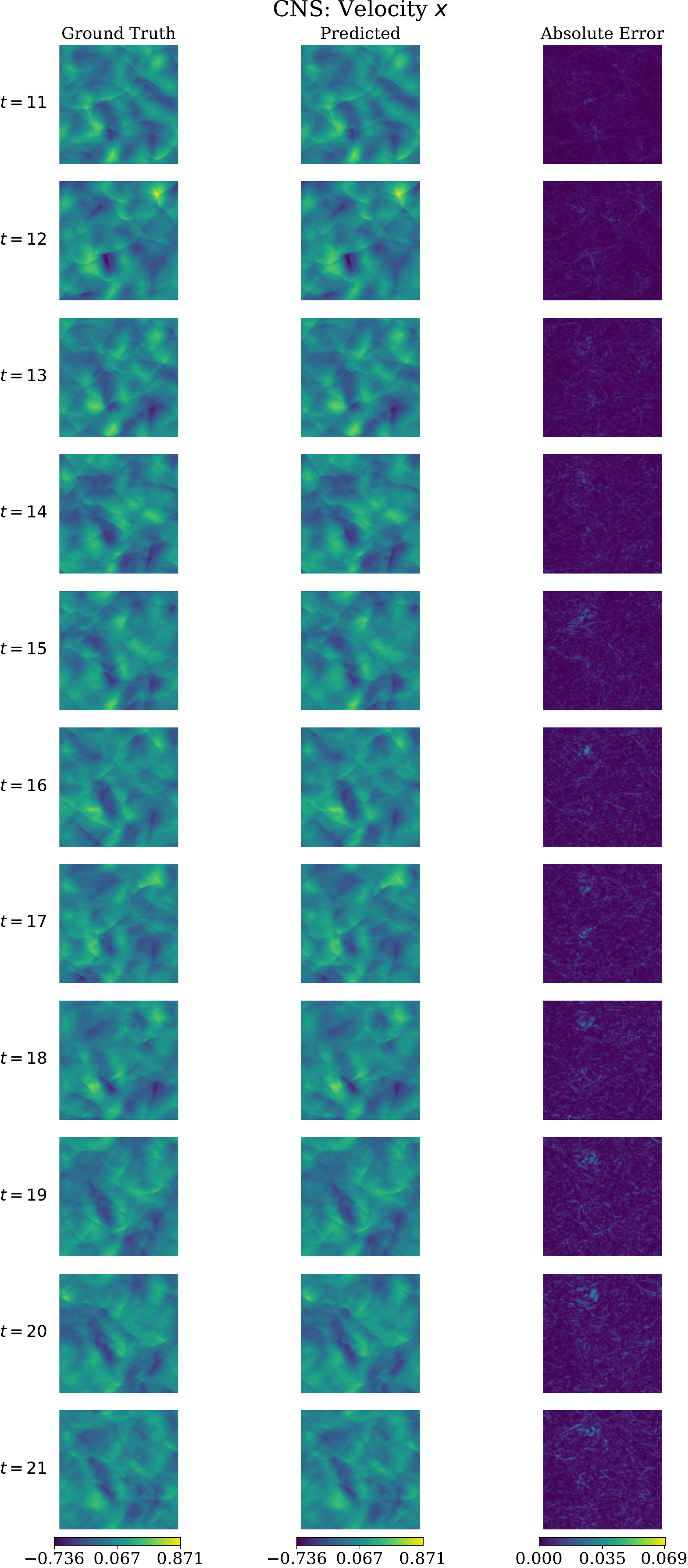}
    \caption{Compressible Navier-Stokes velocity $x$ component.}
    \label{fig:CNSvelx_pred}
\end{figure}

\begin{figure}[t]
    \centering
    \includegraphics[width=0.7\linewidth]{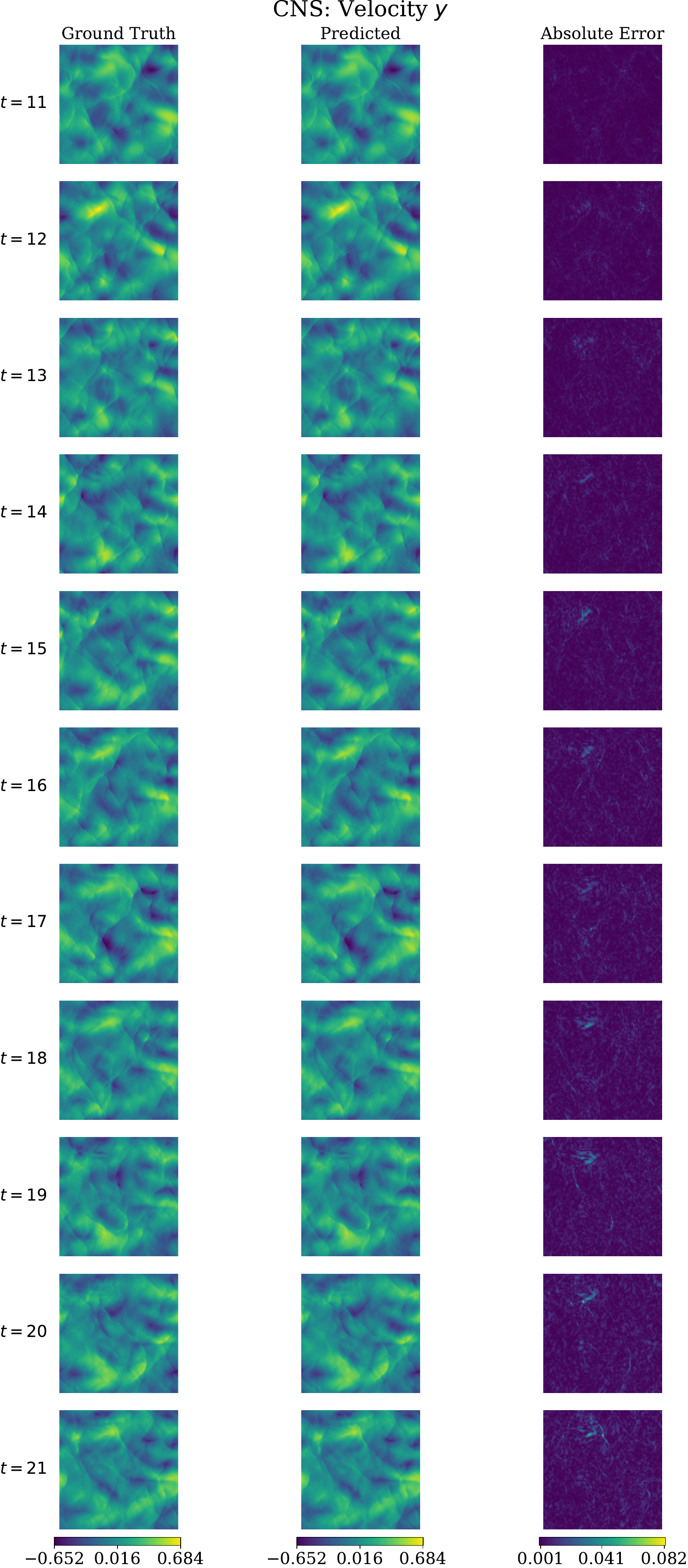}
    \caption{Compressible Navier-Stokes velocity $y$ component.}
    \label{fig:CNSvely_pred}
\end{figure}

\begin{figure}[t]
    \centering
    \includegraphics[width=\linewidth]{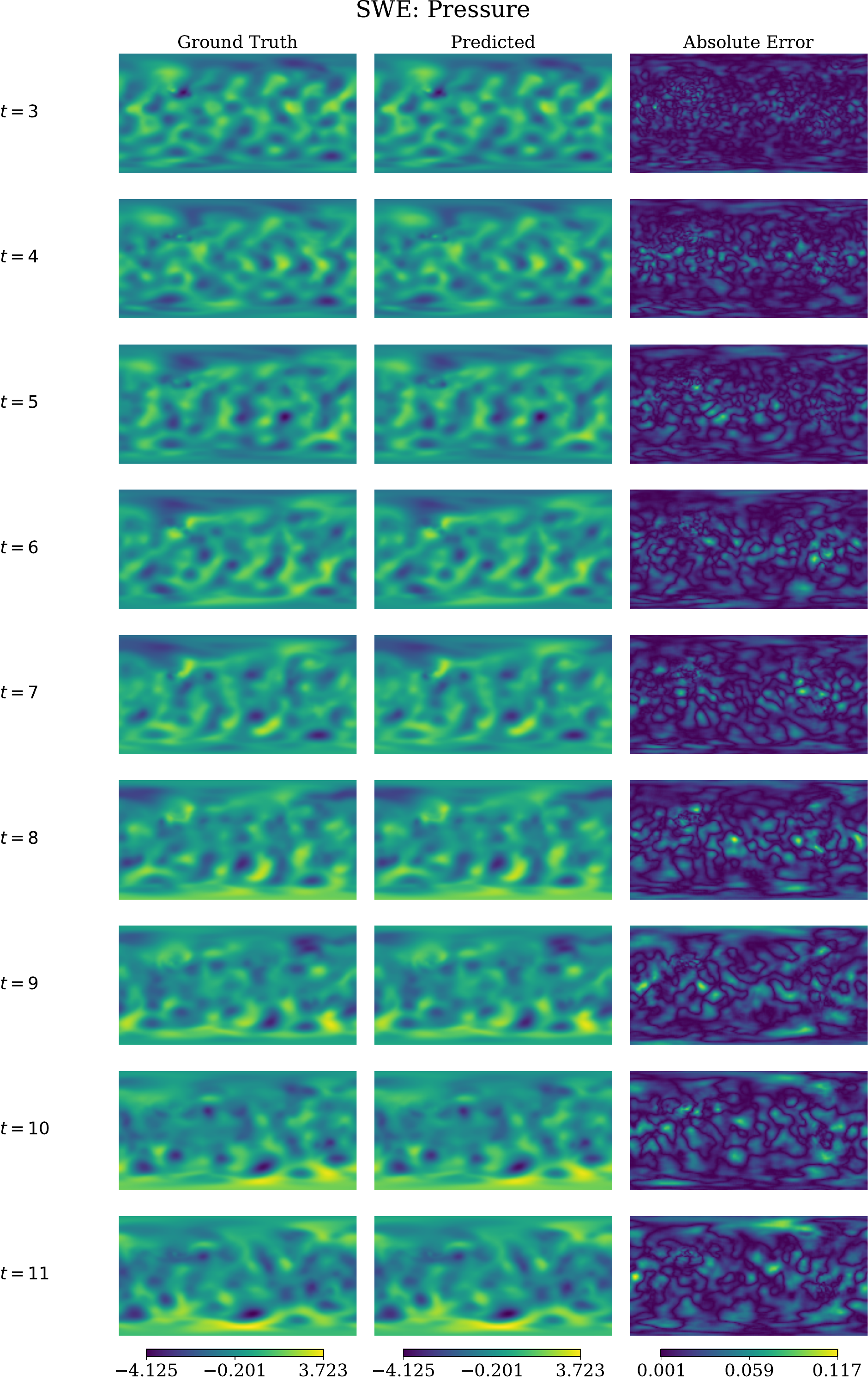}
        \caption{Shallow water equations pressure field.}
    \label{fig:SWEpressure_pred}
\end{figure}

\begin{figure}[t]
    \centering
    \includegraphics[width=\linewidth]{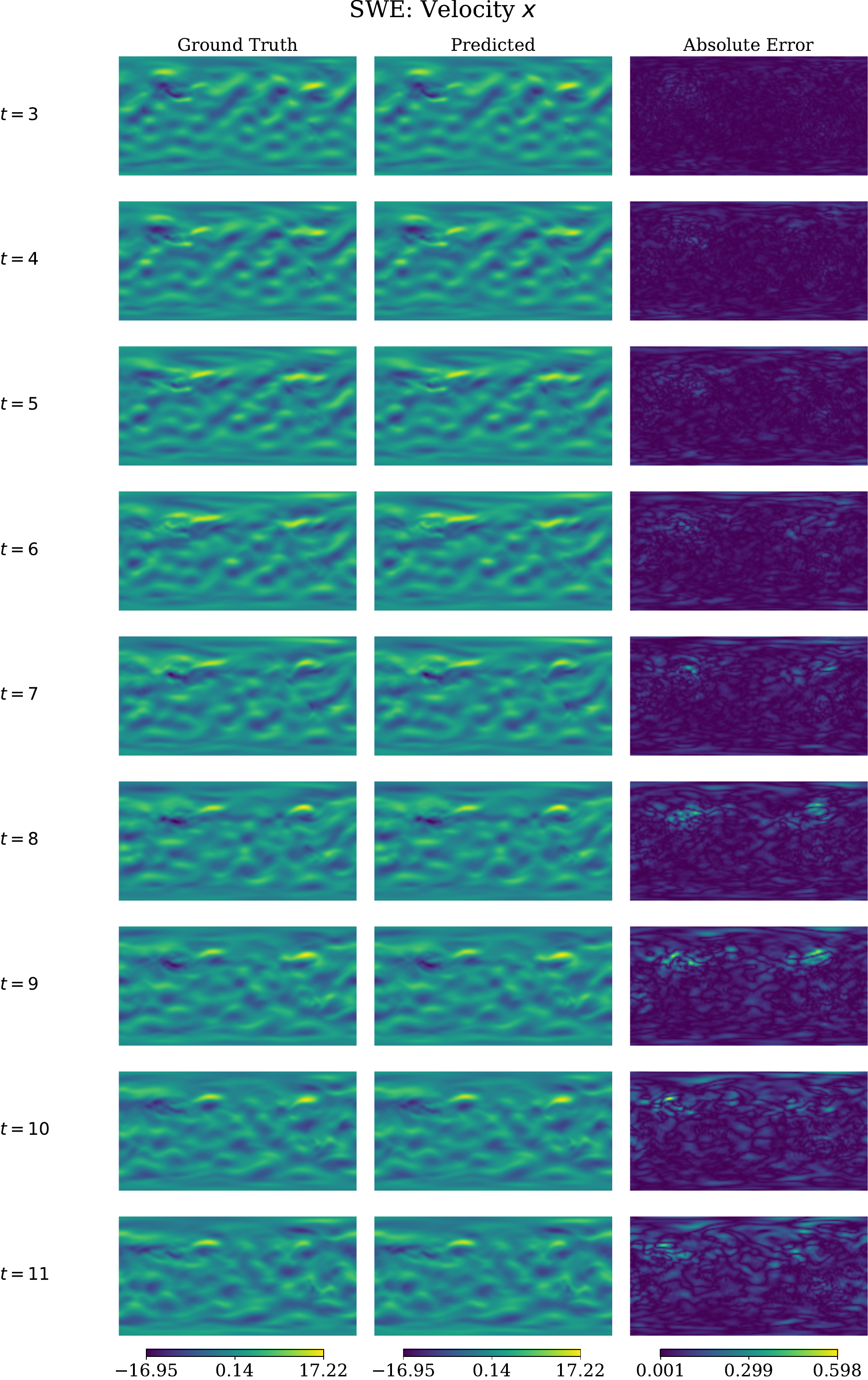}
    \caption{Shallow water equations velocity $x$ component.}
    \label{fig:SWEvelx_pred}
\end{figure}

\begin{figure}[t]
    \centering
    \includegraphics[width=\linewidth]{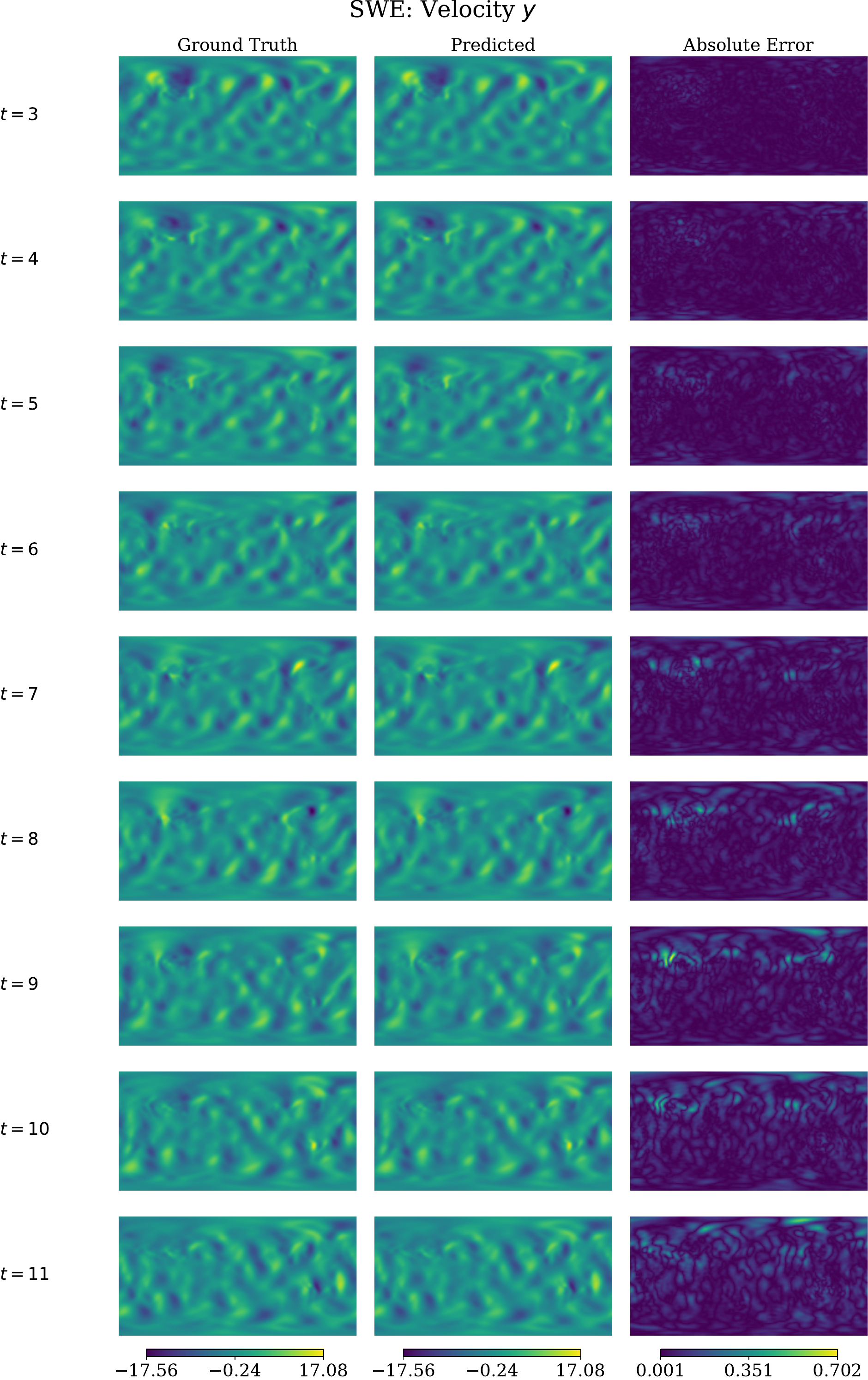}
    \caption{Shallow water equations velocity $y$ component.}
    \label{fig:SWEvely_pred}
\end{figure}

Here, we visualize rollout predictions from \textsc{SineNet-8} on INS (Figures~\ref{fig:INSdensity_pred}-\ref{fig:INSvely_pred}), CNS (Figures~\ref{fig:CNSdensity_pred}-\ref{fig:CNSvely_pred}), and SWE (Figures~\ref{fig:SWEpressure_pred}-\ref{fig:SWEvely_pred}) on a randomly selected trajectory from each test set. In each, we show the ground truth field in the left column, the predicted field in the middle column, and the absolute error in the right column.

% \subsection{INS}\label{app:INS}

% \subsection{CNS}\label{app:CNS}

% \subsection{SWE}\label{app:SWE}

\end{document}